\documentclass[11pt]{article}

\PassOptionsToPackage{switch}{lineno}

\newcommand{\benchmarkname}{CapQuiz}
\newcommand{\benchmarkscore}{CapF1}
\newcommand{\benchmarkprec}{CapP}
\newcommand{\benchmarkrecall}{CapR}
\newcommand{\numvideo}{1,204}
\newcommand{\avgdura}{34.69}
\newcommand{\numqa}{23,632}
\newcommand{\avgqa}{19.63}

\usepackage{booktabs} 
\usepackage{multirow} 
\usepackage{tabularx}  
\usepackage{array}
\usepackage{adjustbox}
\usepackage{enumitem}
\usepackage{colortbl} 
\usepackage{subcaption} 
\usepackage{cuted}
\usepackage{graphicx} 
\usepackage{tikz}
\usetikzlibrary{calc, math, fadings}
\usepackage[most]{tcolorbox}
\tcbuselibrary{skins, breakable} 
\usepackage{xcolor}
\usepackage{pifont}
\usepackage{fvextra}
\newcommand{\cmark}{\ding{51}} 
\newcommand{\xmark}{\ding{55}} 
\newcommand{\best}[1]{\textbf{#1}}
\newcommand{\second}[1]{\underline{#1}}

\newcommand{\red}[1]{\textcolor{red}{\ensuremath{#1}}}
\newcommand{\green}[1]{\textcolor{green!50!black}{\ensuremath{#1}}}

\definecolor{colKnowledge}{HTML}{3498DB} 
\definecolor{colEveryday}{HTML}{F39C12}  
\definecolor{colCreativity}{HTML}{E74C3C} 
\definecolor{colCivics}{HTML}{2ECC71}    
\definecolor{colFunction}{HTML}{9B59B6}  
\definecolor{colPerception}{HTML}{16A085} 
\definecolor{colReasoning}{HTML}{E91E63} 

\definecolor{colSource}{HTML}{34495E} 
\definecolor{colOption}{HTML}{95A5A6} 

\definecolor{colDurShort}{HTML}{66C2A5}
\definecolor{colDurMed}{HTML}{FC8D62}
\definecolor{colDurLong}{HTML}{8DA0CB}

\usepackage[final]{acl}

\usepackage{times}
\usepackage{latexsym}

\usepackage[T1]{fontenc}

\usepackage[utf8]{inputenc}

\usepackage{microtype}

\usepackage{inconsolata}

\usepackage{graphicx}

\title{Putting Captions to the Test: Evaluating Video Caption Quality through Multiple-Choice Question Answering}

\author{
 \textbf{Zizhen Wang},
 \textbf{Bo Feng},
 \textbf{Zhengfeng Lai}\thanks{Work done at Apple.},
 \textbf{Shiyu Li}
\\
 \textbf{Yang Lu},
 \textbf{Meng Cao},
 \textbf{Ping Huang},
 \textbf{Simon Wang}
\\
 Apple
\\
 \small{\{wang\_zizhen, bfeng2, jeff\_lai, shiyu\_li, yanglu, mengcao, huang\_ping, simon\_wang2\}@apple.com}
}

\begin{document}
\maketitle

\begin{abstract}

Evaluating video captioning remains a critical challenge for Visual Large Language Models (VLLMs).
Existing metrics primarily rely on matching generated text against ground-truth references. 
This paradigm suffers from the ``one-to-many'' nature of video description, where high-quality captions are often penalized for lexical mismatches or valid shifts in visual focus.
Furthermore, such assessments are typically one-dimensional, failing to provide a fine-grained analysis of caption quality.
To address this, we redefine caption quality via information fidelity: \textit{A caption must maximize the coverage of salient visual information while ensuring strict factuality.}
We introduce {\benchmarkname}, a novel reference-free benchmark that assesses captions based on their utility in answering human-verified, fine-grained, multiple-choice questions derived from the video. 
{\benchmarkname} features a hierarchical taxonomy of 10 question types (spanning \textit{Descriptive} and \textit{Inferential} categories) across 24 diverse video domains. 
We further formulate {\benchmarkscore}, a composite metric that synthesizes {\benchmarkprec} (measuring factuality) and {\benchmarkrecall} (measuring coverage).
Extensive experiments demonstrate that {\benchmarkname} correlates significantly better with human judgments than existing metrics and offers interpretable insights into model performance. 

\end{abstract}

\section{Introduction}

\begin{figure*}[h]
    \centering  
    \includegraphics[width=1.0\textwidth]{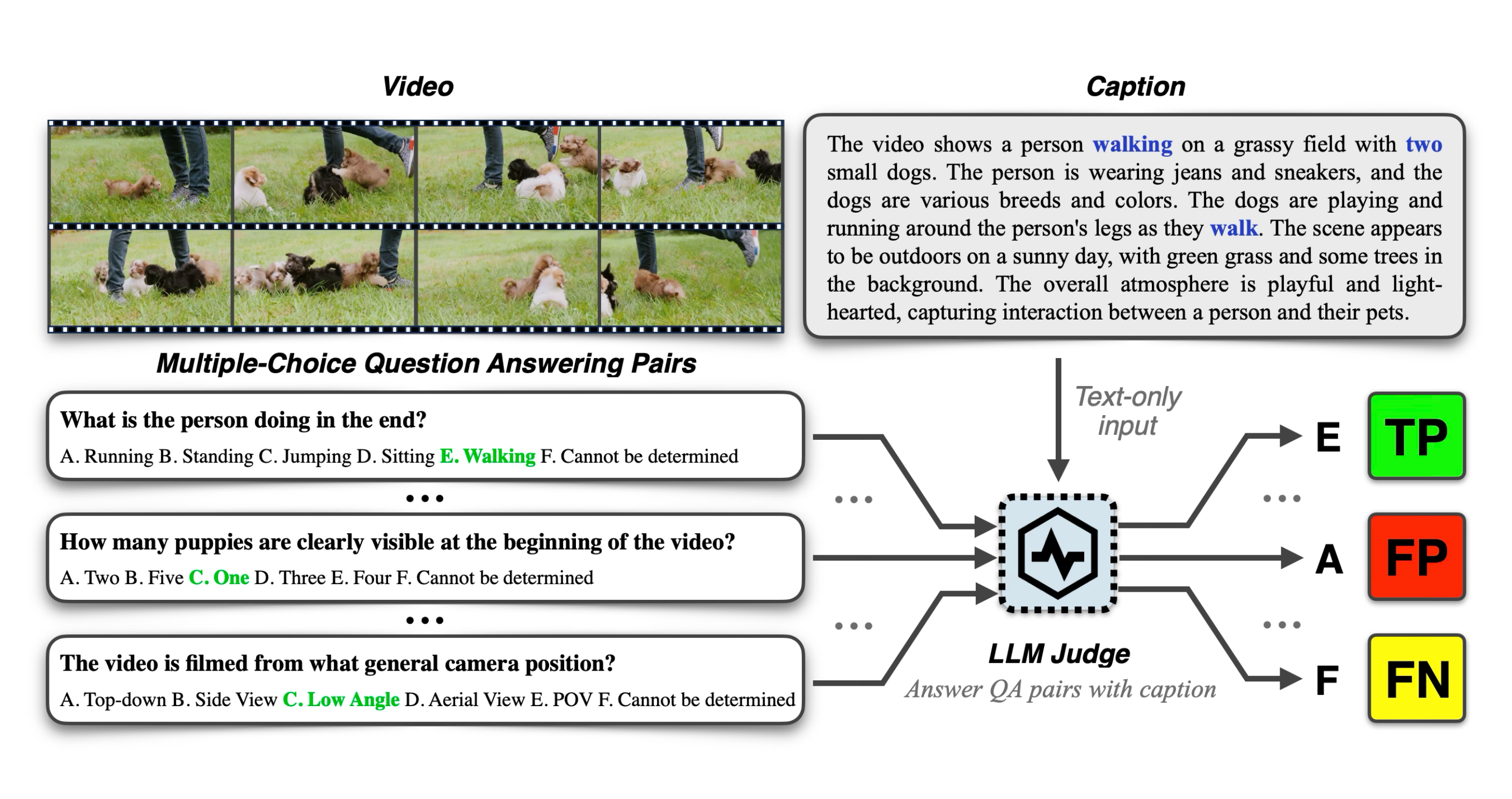} 
    \vspace{-12mm}
    \caption{Overview of the {\benchmarkname} evaluation pipeline. We assess caption quality by leveraging an LLM to answer human-verified fine-grained multiple-choice questions based solely on the generated caption. The green options denote the ground-truth answers, while blue text highlights the key evidence within the caption.} 
    \label{fig:evaluation} 
    \vspace{-4mm}
\end{figure*}

Recent advancements in Visual Large Language Models (VLLMs) have significantly improved the performance of vision-language tasks.
Among these, video captioning remains a fundamental challenge, requiring models to perceive temporal visual dynamics and synthesize them into coherent natural language.
This capability is essential for various downstream applications, such as video retrieval, content indexing, and accessibility tools for the visually impaired.
As VLLMs become more capable of generating detailed and lengthy descriptions, the need for a robust benchmark to assess their quality has become increasingly critical.

The dominant evaluation paradigm remains text-based comparison with ground-truth references, relying on n-gram overlap metrics like BLEU \citep{papineni2002bleu} and CIDEr \citep{vedantam2015cider}, or semantic judges like BERTScore \citep{zhang2019bertscore} and LLM-based evaluators \citep{liu2023g, fu2024gptscore}.
However, these methods fundamentally suffer from the ``one-to-many'' nature of video description, often penalizing valid captions that diverge lexically from the ground truth.
To address this, cross-modal metrics have emerged to incorporate visual grounding into the evaluation loop.
Approaches range from leveraging matching models and scene graphs \citep{jiang2019tiger, wang2021faier} to utilizing pre-trained vision-language embeddings \citep{hessel2021clipscore, hu2023infometic}.
Parallel to this, a distinct research strand has shifted toward question-based evaluation, such as VDC \citep{chai2024auroracap} and VCapsBench \citep{zhang2025vcapsbench}, which probe caption fidelity through question answering.
Crucially, however, neither paradigm successfully provides a fine-grained, diagnostic analysis capable of effectively decoupling factual trustworthiness from information coverage.

To address these limitations, we propose evaluating captions based on information fidelity.
Fundamentally, a high-quality caption should act as an effective textual surrogate for the video, preserving the key visual details accurately.
The core criterion thus becomes: \textit{can a user correctly answer fine-grained questions about the video solely by reading the caption?}
This reference-free approach encourages the model to maximize the coverage of salient information while ensuring factuality regarding the source video.

Guided by this philosophy, we introduce {\benchmarkname}, a benchmark that quantifies caption quality through human-verified, fine-grained, multiple-choice question answering, as shown in Figure \ref{fig:evaluation}.
Unlike binary verification, the multiple-choice format forces the model to discriminate between correct details and plausible distractors, offering a more rigorous test of fine-grained understanding.
{\benchmarkname} is built upon a hierarchical taxonomy comprising 10 specific question types, categorized under \textit{Descriptive} (e.g., entity, attribute) and \textit{Inferential} (e.g., relation, causality).
This structure enables a multi-faceted analysis of model capabilities across 24 diverse video domains.
We categorize the QA results into True Positives (TP), False Positives (FP), and False Negatives (FN). 
Building upon this taxonomy, we introduce {\benchmarkprec} and {\benchmarkrecall} to quantify factuality and coverage, respectively, and aggregate them into a unified metric, {\benchmarkscore}.

Our main contributions are summarized as follows:

\begin{itemize}

\item 
We propose {\benchmarkname}, a novel reference-free benchmark grounded in the principle of information fidelity.
By utilizing human-verified, fine-grained multiple-choice questions with plausible distractors, it robustly assesses caption quality in terms of factuality ({\benchmarkprec}), coverage ({\benchmarkrecall}), and the unified metric ({\benchmarkscore}).

\item 
We release a comprehensive benchmark comprising {\numvideo} videos spanning 24 diverse domains. 
It features {\numqa} human-verified multiple-choice question-answer pairs, organized into a rigorous taxonomy of 10 specific types under broader \textit{Descriptive} and \textit{Inferential} categories.

\item 
Extensive experiments demonstrate that {\benchmarkname} achieves superior alignment with human preferences. 
Our fine-grained analysis further reveals VLLM disparities and diagnostic insights missed by traditional one-dimensional metrics.

\end{itemize}

\begin{figure*}[h]
    \centering
    \includegraphics[width=1.0\textwidth]{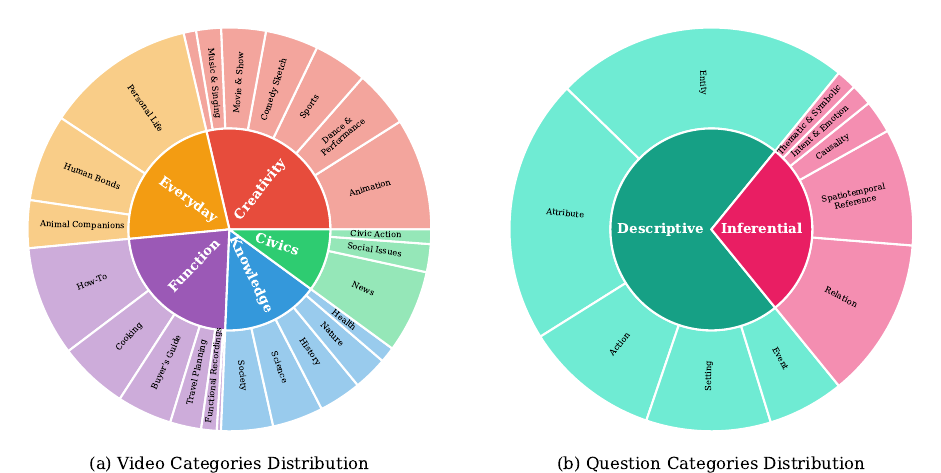}
    \vspace{-4mm} 
    \caption{Hierarchical statistics of the proposed {\benchmarkname}.}
    \label{fig:taxonomy}
    \vspace{-4mm} 
\end{figure*}

\section{Related Works}

\textbf{Text-based Evaluation} relies on comparing candidates against human-authored reference captions.
Traditional n-gram metrics (e.g. BLEU \citep{papineni2002bleu}, ROUGE \citep{lin2004rouge}) focus on surface-level lexical overlap, while METEOR \citep{banerjee2005meteor} incorporates synonymy via WordNet \citep{miller1995wordnet}.
CIDEr \citep{vedantam2015cider}, specifically designed for image captioning, computes cosine similarity using TF-IDF weighting to emphasize distinctive terms.
To capture structural semantics, SPICE \citep{anderson2016spice} parses captions into scene graphs composed of objects, attributes, and relationships.
However, these matching paradigms suffer from the ``one-to-many'' nature of video description, penalizing valid captions that deviate lexically from the ground truth.
Recent semantic metrics like BERTScore \citep{zhang2019bertscore}, BERTHA \citep{lebron2022bertha} and LLM-based judges \citep{liu2023g, fu2024gptscore} move beyond exact matches but remain inherently reference-dependent.
They are constrained by the limited coverage of ground-truth annotations and suffer from reference bias, where high-quality captions are undervalued simply for describing valid visual details absent in the specific reference texts.


\textbf{Cross-modal Grounded Evaluation} mitigates reference reliance by directly incorporating visual information into the evaluation loop.
Early approaches leveraged image-text matching models \citep{lee2018stacked, jiang2019tiger} or scene graphs \citep{wang2020cross, wang2021faier} to score fidelity.
With the advent of large-scale pre-training, CLIPScore \citep{hessel2021clipscore} has become a de facto standard, computing the cosine similarity in a shared semantic space provided by models like CLIP \citep{radford2021learning}.
InfoMetIC \citep{hu2023infometic} builds on VLMs to provide both coarse-grained and token-level quality scores.
Despite their popularity, these metrics typically yield a global similarity score that treats the caption as a ``bag of words'', often failing to distinguish fine-grained semantic nuances such as object relations or action directionality.
Crucially, they lack explicit modeling of temporal dynamics and logical reasoning, making them less effective in diagnosing whether a model truly understands the complex events unfolding in a video.


\textbf{Question-based Evaluation} assesses caption quality via information fidelity.
While early works like QACE \citep{lee2021qace}, VQAScore \citep{lin2024evaluating} and CaptionQA \citep{yang2025captionqa} utilize visual question answering to verify consistency, they are primarily tailored for static images.
In the video domain, Dream1K \citep{wang2024tarsier} compares answers derived from candidates against those from references; however, this reintroduces the ``one-to-many'' limitation where valid but non-overlapping information is penalized.
To enable reference-free evaluation, VDC \citep{chai2024auroracap}, QEVA \citep{jung2025qeva} and VCapsBench \citep{zhang2025vcapsbench} introduce QA sets to bypass this issue.
Nevertheless, such binary or open-ended formats are susceptible to random guessing or instability, lacking the plausible distractors necessary for fine-grained discrimination.
Critically, most existing QA metrics predominantly verify factual correctness, largely overlooking whether the caption provides comprehensive coverage of salient events.
In contrast, our work explicitly decomposes caption quality into factual trustworthiness and information coverage, treating question answerability not merely as a verification signal but as the primary evaluation objective.


\begin{figure*}[t]
    \centering
    \vspace{-4mm}
    \includegraphics[width=1.0\textwidth]{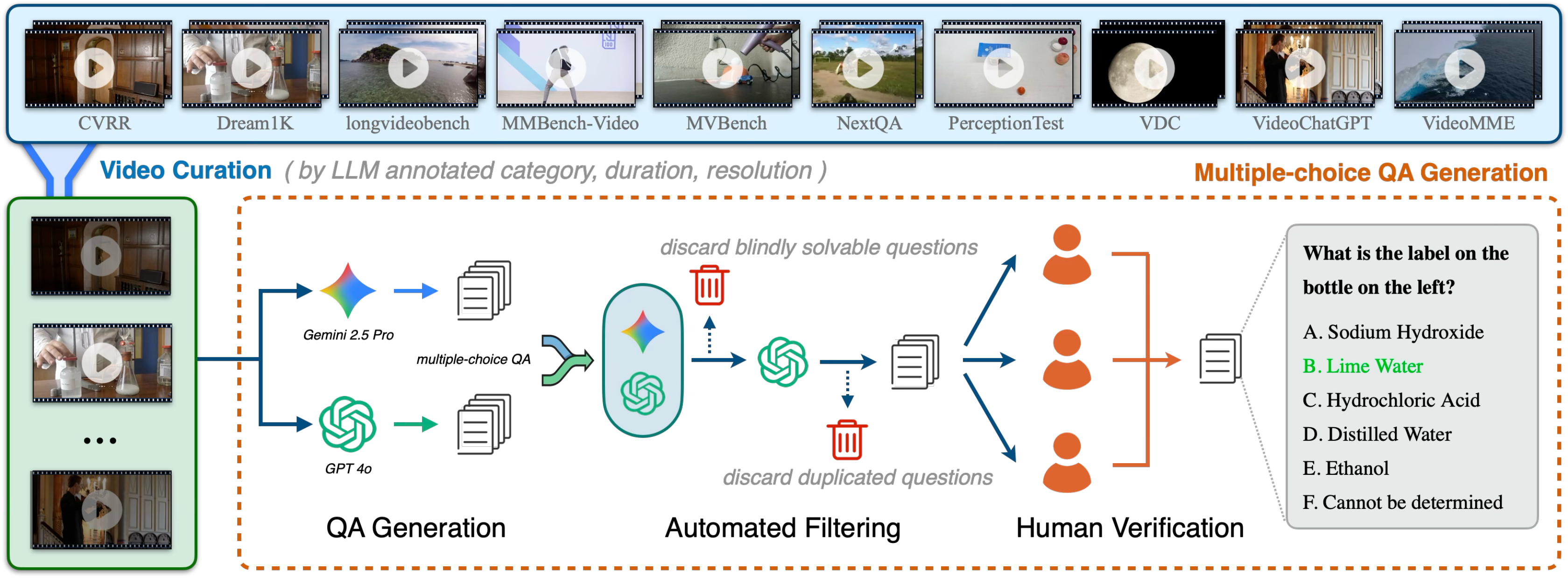}
    \caption{The construction pipeline of \benchmarkname.}
    \vspace{-4mm}
    \label{fig:pipeline}
\end{figure*}

\section{The {\benchmarkname} Benchmark}

In this section, we introduce the methodology behind {\benchmarkname}, a reference-free benchmark designed to evaluate video caption quality through human-verified, fine-grained multiple-choice question answering.

\subsection{Hierarchical Taxonomy Design}

To ensure a holistic assessment of VLLM caption capabilities, we ground {\benchmarkname} in a rigorous two-level taxonomy governing both visual domains and the probing question types.
As shown in Figure \ref{fig:taxonomy}(a), we curate videos across 5 super-categories (e.g., \textit{Knowledge}, \textit{Everyday}) branching into 24 fine-grained sub-categories.
This stratification maximizes semantic diversity, ranging from dynamic events in \textit{Sports} to information-dense scenes in \textit{News}, ensuring models are tested against distinct visual distributions and temporal dynamics.
Complementing this visual breadth, our question taxonomy (Figure \ref{fig:taxonomy}(b)) probes information fidelity across two cognitive dimensions: \textit{Descriptive} that focusing on visual grounding tasks like \textit{Entity} and \textit{Action}, and \textit{Inferential} that targeting higher-order logic such as \textit{Relation} and \textit{Causality}.
By organizing 10 specific question types under these categories, we effectively decouple basic recognition from complex interpretation, enabling fine-grained diagnostic analysis.
Detailed definitions of both taxonomies are provided in Appendix \ref{app:taxonomy}.

\subsection{Benchmark Construction}

Figure \ref{fig:pipeline} shows the construction pipeline of the benchmark.

\subsubsection{Video Curation}

Preventing data contamination was a paramount priority in our construction process.
To minimize the risk of training set leakage, we strictly sourced candidate videos from the test or held-out validation splits of 10 public benchmarks, including VideoMME \citep{fu2025video}, VideoChatGPT \citep{maaz2024video}, NextQA \citep{xiao2021next}, MVBench \citep{li2024mvbench}, MMBench-Video \citep{fang2024mmbench}, CVRR \citep{khattak2025good}, PerceptionTest \citep{patraucean2023perception}, longvideobench \citep{wu2024longvideobench}, VDC \citep{chai2024auroracap} and Dream1K \citep{wang2024tarsier}.

We extracted essential metadata (i.e., resolution, duration) and employed \texttt{Gemini-2.5-Pro} \citep{comanici2025gemini} to annotate video categories, which guided the subsequent selection process.
We adopted a duration ratio of approximately \textit{6:3:1} for short (0--30s), medium (30--60s), and long (60--120s) videos.
While ensuring ample coverage of short-form clips, this distribution also incorporates narratively rich content through longer videos.

\begin{table*}[t]
    \centering
    \small 
    \resizebox{\textwidth}{!}{ 
    \begin{tabular}{lcccccccc}
        \toprule
        \textbf{Benchmark} & \textbf{Reference-free} & \textbf{Human Verified} & \textbf{QA Format} &\textbf{\# Videos} & \textbf{Avg. Duration (s)} &  \textbf{Avg. Q/V} \\
        \midrule
        MSVD \citeyearpar{chen2011collecting} & \xmark & - & -  & 1,970 & 9.65 & - \\
        MSR-VTT \citeyearpar{xu2016msr} & \xmark & - & - & 10,000 & 15.01 & - \\
        ActivityNet Captions \citeyearpar{krishna2017dense} & \xmark & - & - & 9,802 & 118.21 & - \\
        VATEX \citeyearpar{wang2019vatex} & \xmark & - & - & 4,478 & 144.78 & - \\
        Dream1K \citeyearpar{wang2024tarsier} & \xmark & - & - & 1,000 & 8.87 & - \\
        VDC \citeyearpar{chai2024auroracap} & \cmark & \xmark & Open-Ended & 1,027 & 28.18 & 94.35 \\
        VCapsBench \citeyearpar{zhang2025vcapsbench} & \cmark & \cmark & Yes/No & 5,677 & 9.79 & 18.38 \\
        \midrule
        \textbf{\benchmarkname} & \cmark & \cmark  & Multiple-Choice & {\numvideo} & {\avgdura} & {\avgqa} \\
        \bottomrule
    \end{tabular}
    }
    \caption{\textbf{Comparison of {\benchmarkname} with existing video captioning benchmarks.} 
    Our benchmark provides human-verified fine-grained multiple-choice question answering pairs designed for caption information fidelity evaluation.
    Avg. Q/V indicates Average number of questions per video. 
    }
    \vspace{-4mm}
    \label{tab:benchmark_comparison}
\end{table*}

\begin{figure*}[t]
    \centering
    \includegraphics[width=1.0\textwidth]{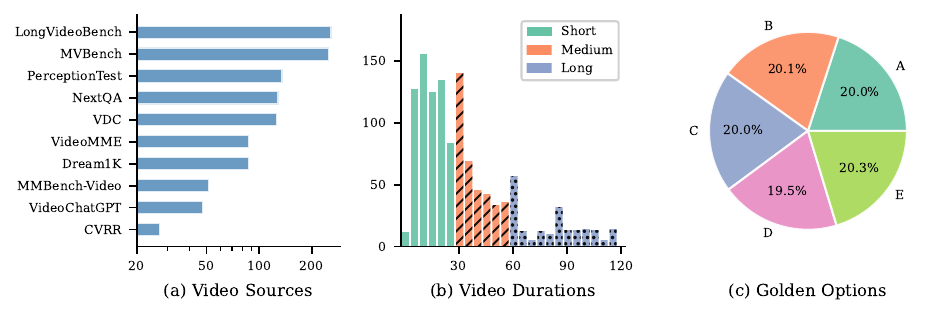}
    \vspace{-8mm}
    \caption{Detailed statistics of the proposed {\benchmarkname}.}
    \label{fig:stat}
    \vspace{-4mm} 
\end{figure*}

\subsubsection{Multiple-Choice QA Generation}

We implemented a rigorous \textit{Over-generate then Filter} pipeline to construct the multiple-choice question-answer set.
In the generation phase, utilizing \texttt{Gemini-2.5-Pro} and \texttt{GPT-4o} \citep{hurst2024gpt}, we produced diverse question-answer pairs rooted in our hierarchical taxonomy.
Specifically, we prompted the models with the raw video, the specific definition of the target question type, and at least five few-shot reference examples. 
The models were instructed to generate a list of candidate QA pairs, where each pair comprises a distinct question body and five answer options.

Subsequently, these candidates underwent a strictly controlled filtration phase, beginning with an automated stage designed to ensure validity and information density. 
We first addressed language bias via a \textit{Blind Solvability Check}. 
In this step, QA pairs were fed to LLMs (i.e., \texttt{Gemini-2.5-Pro} and \texttt{GPT-4.1}) in a video-blind setting, with option orders shuffled across three independent trials for each LLM. 
A question was deemed \textit{Blindly Solvable} if both models answered it correctly in at least two out of the three trials, suggesting the answer could be inferred solely from textual patterns without visual context. 
Following this, we performed semantic de-duplication to eliminate redundancy. 
We utilized \texttt{GPT-4.1} to analyze semantic similarity and cluster related questions. 
Within each cluster, we retained only the single most challenging instance, quantified as the one yielding the lowest accuracy during the blind pass, thereby maximizing the discriminative power of the dataset.

Finally, the surviving candidates advanced to the human stage for rigorous verification. 
During this process, annotators were allowed and explicitly instructed to re-watch the video as needed. 
Each QA pair was evaluated by at least three annotators based on three strict criteria: 

\begin{itemize}
    \item \textbf{Visual Relevance:} A question is considered valid only if it is strictly grounded in the video.
    \item \textbf{Factual Correctness:} Each question must have exactly one correct answer that is objectively verifiable through visual evidence in the video. 
    \item \textbf{Question difficulty:} To ensure the benchmark evaluates fine-grained visual understanding, we require that at least one incorrect option be a ``hard negative.'' A hard negative is an option that appears plausible but is factually incorrect, requiring careful inspection of the video details to rule out.
\end{itemize}

To validate the reliability of human review process, we calculated the inter-annotator agreement on a subset of the data, achieving a Gwet's AC1 \citep{gwet2001handbook} of 0.92, which indicates the high consistency and quality of our ground-truth annotations. 
The prompts are listed in Appendix \ref{app:prompt}.

\subsection{Dataset Statistics}

As shown in Table \ref{tab:benchmark_comparison}, {\benchmarkname} contains {\numvideo} videos, with an average of {\avgqa} QA pairs per video.
Unlike reference-based benchmarks (e.g., MSR-VTT), {\benchmarkname} adopts a reference-free paradigm to circumvent the ``one-to-many'' constraints of text matching, thereby directly assessing information fidelity. 
Its human-verified multiple-choice format ensures deterministic and reliable evaluation, superior to the reference-free attempts limited by unverified generation or binary tasks.
Furthermore, featuring a hierarchical taxonomy of 24 video domains and 10 question types, {\benchmarkname} enables a more nuanced diagnosis of model capabilities than prior coarse-grained datasets.

As visually detailed in Figure \ref{fig:taxonomy}, our benchmark is structured around a sophisticated hierarchical video and question taxonomy. 
This design meticulously balances visual richness with linguistic complexity, ensuring that the benchmark covers a wide spectrum of semantic granularity, from coarse-grained object recognition to fine-grained reasoning.
Furthermore, the source distribution presented in Figure \ref{fig:stat}(a) demonstrates that we aggregate video data from a highly heterogeneous array of sources, while minimizing the risk of training set leakage. 
This strategy is intended to maximize visual diversity and domain coverage, thereby testing the generalization ability of models across different visual styles.
In terms of video duration (Figure \ref{fig:stat}(b)), while the benchmark is primarily anchored in short-form videos ($<30$s) to capture atomic events, we deliberately maintain a substantial proportion of medium- and long-form content. 
This diverse duration distribution serves as a rigorous test for models' temporal reasoning capabilities and their ability to model long-range dependencies.
Finally, to ensure fair evaluation, the answer options are strictly uniformly distributed as shown in Figure \ref{fig:stat}(c). 
This balance is critical for mitigating potential position bias and prevents models from bypassing genuine understanding by exploiting statistical shortcuts or spurious correlations.

\subsection{Evaluation Methodology}

To evaluate the quality of video caption $C$, we define a question set $\mathcal{Q}$ derived from the video.
Each question comprises a stem $q_i$ and $5$ options $\mathcal{O}_i = \{o_{i,1}, o_{i,2}, \dots, o_{i,5}\}$, where $o_{i, gt \in [1, \dots, 5]}$ is the correct option. 
We introduce an extra \textit{``Cannot be determined''} option $o_{unk}$ to form the evaluation space $\mathcal{O}'_i = \mathcal{O}_i \cup \{o_{unk}\}$, allowing the \texttt{Judge} to explicitly signal uncertainty due to missing information.
Given the caption $C$ and question $q_i$, the \texttt{judge} selects the most likely option $\hat{o}_i$ as the answer from options $\mathcal{O}'_i$.
The outcomes are categorized as:
\begin{itemize}
\item \textbf{True Positive (TP)} ($\hat{o}_i = o_{i, gt}$): The caption contains the correct visual information, enabling the \texttt{judge} to select the ground-truth option.
\item \textbf{False Negative (FN)} ($\hat{o}_i = o_{unk}$): The caption lacks the necessary information to answer the question. This results in an \textit{omission}, forcing the \texttt{judge} to choose "Cannot be determined."
\item \textbf{False Positive (FP)} ($\hat{o}_i \neq o_{i, gt} \land \hat{o}_i \neq o_{unk}$): The caption contains incorrect or misleading details consistent with a wrong option. This reflects \textit{hallucination}, leading the \texttt{judge} to a specific incorrect answer.
\end{itemize}



Based on the categorization outcomes, we propose three metrics to quantify caption quality:

\begin{itemize}
\item \textbf{Factuality (\benchmarkprec).} This metric measures the precision of determinate answers, penalizing hallucinations (FP) while disregarding uncertainty (FN).
\begin{equation*}
\benchmarkprec = \frac{TP}{TP + FP}
\end{equation*}

\item \textbf{Coverage (\benchmarkrecall).} This metric evaluates the completeness of the caption by measuring the proportion of correctly retrieved details against the total number of questions $N$.
\begin{equation*}
    \benchmarkrecall = \frac{TP}{N}
\end{equation*}

\item \textbf{Overall (\benchmarkscore).} To provide a holistic assessment, we compute the harmonic mean of factuality and coverage:
\begin{equation*}
    \benchmarkscore = \frac{2 \cdot \benchmarkprec \cdot \benchmarkrecall}{\benchmarkprec + \benchmarkrecall}
\end{equation*}
\end{itemize}

To ensure reproducibility and minimize prior knowledge bias, we employ \texttt{GPT-4.1} as the \texttt{judge} in our experiments, instructing it to strictly ground its answers in the provided caption $C$.

\section{Experiments}

\subsection{Alignment with Human Judgments}

\begin{table}[t]
    \centering
    \resizebox{\columnwidth}{!}{
        \begin{tabular}{llcc}
            \toprule
            \textbf{Correlation} & \textbf{Dimension} & \textbf{VLLM-as-a-Judge} & \textbf{Ours} \\
            \midrule
            \multirow{3}{*}{\textbf{Spearman ($\rho$)}} 
            & Factuality & $0.556\;_{[0.421, 0.674]}$ & $\mathbf{0.690}\;_{[0.593, 0.765]}$ \\
            & Coverage   & $0.579\;_{[0.456, 0.701]}$ & $\mathbf{0.663}\;_{[0.562, 0.745]}$ \\
            & Overall    & $0.522\;_{[0.389, 0.650]}$ & $\mathbf{0.643}\;_{[0.542, 0.728]}$ \\
            \midrule
            \multirow{3}{*}{\textbf{Kendall ($\tau$)}} 
            & Factuality & $0.500\;_{[0.378, 0.607]}$ & $\mathbf{0.581}\;_{[0.498, 0.649]}$ \\
            & Coverage   & $0.515\;_{[0.406, 0.624]}$ & $\mathbf{0.551}\;_{[0.463, 0.624]}$ \\
            & Overall    & $0.463\;_{[0.342, 0.577]}$ & $\mathbf{0.533}\;_{[0.447, 0.608]}$ \\
            \bottomrule
        \end{tabular}
    }
    \caption{\textbf{Correlation with human judgments.} Spearman ($\rho$) and Kendall ($\tau$) correlations are reported for \textit{Factuality}, \textit{Coverage}, and \textit{Overall Quality}, with 95\% bootstrap confidence intervals over videos. Our {\benchmarkname} achieves the highest correlation in all six settings. All correlations are significant ($p < 0.01$).}
    \vspace{-4mm}
    \label{tab:human_corr}
\end{table}

\begin{table*}[t]
    \centering
    \small 
    \setlength{\tabcolsep}{4.5pt} 
    
    \begin{tabular}{l ccc ccc ccc}
        \toprule
        \multirow{2}{*}{\textbf{Model}} & \multicolumn{3}{c}{\textbf{Overall}} & \multicolumn{3}{c}{\textbf{Descriptive}} & \multicolumn{3}{c}{\textbf{Inferential}} \\
        \cmidrule(lr){2-4} \cmidrule(lr){5-7} \cmidrule(lr){8-10}
         & F1 & P & R & F1 & P & R & F1 & P & R \\
        \midrule
        
        \multicolumn{10}{l}{\textit{Proprietary Models}} \\
        Gemini-2.5-flash \citeyearpar{comanici2025gemini} & 81.01 & 84.52 & 77.79 & 83.85 & 87.69 & 80.34 & 73.84 & 76.58 & 71.29 \\
        Gemini-2.5-pro \citeyearpar{comanici2025gemini} & \second{81.71} & \second{84.94} & \second{78.72} & \second{84.30} & \second{87.74} & \second{81.11} & \second{75.16} & \second{77.88} & \second{72.63} \\
        GPT-4o-2024-11-20 \citeyearpar{hurst2024gpt} & 71.97 & 78.13 & 66.71 & 74.99 & 82.03 & 69.07 & 64.45 & 68.70 & 60.70 \\
        GPT-4.1-2025-04-14 \citeyearpar{oai2025gpt41} & 76.79 & 81.65 & 72.47 & 79.34 & 84.82 & 74.53 & 70.40 & 73.88 & 67.24 \\
        GPT-5.2-2025-12-11 \citeyearpar{openai2025gpt51} & \best{83.08} & \best{86.31} & \best{80.09} & \best{85.23} & \best{88.74} & \best{81.99} & \best{77.65} & \best{80.22} & \best{75.24} \\
        \midrule
        
        \multicolumn{10}{l}{\textit{Open-source Models}} \\
        AuroraCap-7B \citeyearpar{chai2024auroracap} & 39.86 & 63.00 & 29.16 & 43.86 & 68.40 & 32.28 & 29.48 & 48.26 & 21.22 \\
        LLaVA-Video-7B \citeyearpar{zhang2024videoinstructiontuningsynthetic} & 63.16 & 73.61 & 55.30 & 66.83 & 78.02 & 58.45 & 53.85 & 62.51 & 47.30 \\
        Tarsier2-7b \citeyearpar{yuan2025tarsier2advancinglargevisionlanguage} & 42.37 & 55.80 & 34.16 & 43.55 & 57.91 & 34.90 & 39.44 & 50.70 & 32.27 \\
        InternVL3.5-8B \citeyearpar{wang2025internvl3} & 45.69 & 67.57 & 34.51 & 47.67 & 72.46 & 35.52 & 40.89 & 56.75 & 31.95 \\
        InternVL3.5-30B-A3B \citeyearpar{wang2025internvl3} & 54.46 & 70.03 & 44.55 & 57.92 & 75.03 & 47.16 & 45.80 & 57.83 & 37.92 \\
        \midrule
        
        \multicolumn{10}{l}{\textit{Qwen3-VL Series} \citeyearpar{bai2025qwen3vltechnicalreport}} \\
        Qwen3-VL-2B & 61.53 & 74.77 & 52.27 & 65.46 & 79.80 & 55.50 & 51.61 & 62.23 & 44.08 \\
        Qwen3-VL-4B & 70.15 & 78.33 & 63.51 & 73.45 & 82.22 & 66.38 & 61.79 & 68.59 & 56.22 \\
        Qwen3-VL-8B & 70.72 & 78.23 & 64.52 & 74.26 & 82.45 & 67.55 & 61.82 & 67.76 & 56.83 \\
        Qwen3-VL-30B-A3B & 68.82 & 78.30 & 61.40 & 71.99 & 82.42 & 63.90 & 60.92 & 68.21 & 55.04 \\
        Qwen3-VL-32B & 77.95 & 82.07 & 74.23 & 80.74 & 85.22 & 76.70 & 70.92 & 74.19 & 67.93 \\
        
        \bottomrule
    \end{tabular}
    
    \caption{\textbf{Main results on {\benchmarkname}}. We report performance across the Overall metric and its two question categories: Descriptive and Inferential. P, R, and F1 correspond to {\benchmarkprec}, {\benchmarkrecall}, and {\benchmarkscore}, respectively. The best performance is marked in \textbf{bold} and the second best is \underline{underlined}.}
    \label{tab:main_results}
    \vspace{-4mm}
\end{table*}

To validate the effectiveness of {\benchmarkname}, we assess the alignment between automated metrics and human judgments using Spearman ($\rho$) \citep{spearman1961general} and Kendall ($\tau$) \citep{kendall1948rank} rank coefficients.
We randomly sampled 200 videos from the benchmark, where the captions were generated by three different models (i.e. \texttt{GPT-4o}, \texttt{InternVL3.5-8B} and \texttt{QWen3-VL-8B}), yielding a total of 600 video-caption pairs.
Human experts and the VLLM-as-a-Judge rated the generated captions on a Likert scale of 1-5 across \textit{Factuality}, \textit{Coverage}, and \textit{Overall Quality}. 
For the VLLM-as-a-Judge, we prompted \texttt{Gemini-2.5-pro} to judge based on the video content (prompt detailed in Appendix \ref{app:prompt}).
For {\benchmarkname}, we map the metric components to the evaluation dimensions: {\benchmarkprec} evaluates \textit{Factuality}, {\benchmarkrecall} measures \textit{Coverage}, and {\benchmarkscore} represents \textit{Overall Quality}.
We report both point estimates and 95\% confidence intervals computed via bootstrap resampling over videos (1,000 resamples).
As shown in Table \ref{tab:human_corr}, {\benchmarkname} consistently achieves higher correlations with human judgments than the VLLM-as-a-Judge baseline across all six settings, and all reported correlations are statistically significant ($p < 0.01$).
These results further indicate that decomposing evaluation into fine-grained QA tasks yields a more reliable and interpretable assessment than direct scoring.

\begin{table*}[t]
    \centering
    \small 
    \setlength{\tabcolsep}{4.5pt} 
    \renewcommand{\arraystretch}{1.5}
    
    \begin{tabular}{l ccc ccc ccc}
        \toprule
        \multirow{2}{*}{\textbf{Model}} & \multicolumn{3}{c}{\textbf{Overall}} & \multicolumn{3}{c}{\textbf{Descriptive}} & \multicolumn{3}{c}{\textbf{Inferential}} \\
        \cmidrule(lr){2-4} \cmidrule(lr){5-7} \cmidrule(lr){8-10}
         & F1 & P & R & F1 & P & R & F1 & P & R \\
        \midrule

        \multicolumn{10}{l}{\textit{Proprietary Models}} \\
        Gemini-2.5-flash & $\substack{ 78.52 \\ \red{-3.07\%}}$ & $\substack{ 82.36 \\ \red{-2.56\%}}$ & $\substack{ 75.02 \\ \red{-3.56\%}}$ & $\substack{ 81.47 \\ \red{-2.84\%}}$ & $\substack{ 85.65 \\ \red{-2.33\%}}$ & $\substack{ 77.68 \\ \red{-3.31\%}}$ & $\substack{ 71.06 \\ \red{-3.76\%}}$ & $\substack{ 74.11 \\ \red{-3.23\%}}$ & $\substack{ 68.26 \\ \red{-4.25\%}}$ \\
        Gemini-2.5-pro & $\substack{ 84.46 \\ \green{+3.37\%}}$ & $\substack{ 86.08 \\ \green{+1.34\%}}$ & $\substack{ 82.89 \\ \green{+5.30\%}}$ & $\substack{ 87.49 \\ \green{+3.78\%}}$ & $\substack{ 89.12 \\ \green{+1.57\%}}$ & $\substack{ 85.92 \\ \green{+5.93\%}}$ & $\substack{ 76.72 \\ \green{+2.08\%}}$ & $\substack{ 78.30 \\ \green{+0.54\%}}$ & $\substack{ 75.20 \\ \green{+3.54\%}}$ \\
        GPT-4o-2024-11-20 & $\substack{ 75.75 \\ \green{+5.25\%}}$ & $\substack{ 79.39 \\ \green{+1.61\%}}$ & $\substack{ 72.43 \\ \green{+8.57\%}}$ & $\substack{ 79.50 \\ \green{+6.01\%}}$ & $\substack{ 83.47 \\ \green{+1.76\%}}$ & $\substack{ 75.89 \\ \green{+9.87\%}}$ & $\substack{ 66.27 \\ \green{+2.82\%}}$ & $\substack{ 69.13 \\ \green{+0.63\%}}$ & $\substack{ 63.64 \\ \green{+4.84\%}}$ \\
        GPT-4.1-2025-04-14 & $\substack{ 80.48 \\ \green{+4.81\%}}$ & $\substack{ 82.30 \\ \green{+0.80\%}}$ & $\substack{ 78.73 \\ \green{+8.64\%}}$ & $\substack{ 83.89 \\ \green{+5.73\%}}$ & $\substack{ 85.82 \\ \green{+1.18\%}}$ & $\substack{ 82.04 \\ \green{+10.08\%}}$ & $\substack{ 71.81 \\ \green{+2.00\%}}$ & $\substack{ 73.37 \\ \red{-0.69\%}}$ & $\substack{ 70.32 \\ \green{+4.58\%}}$ \\
        GPT-5.2-2025-12-11 & $\substack{ 85.14 \\ \green{+2.48\%}}$ & $\substack{ 87.04 \\ \green{+0.85\%}}$ & $\substack{ 83.33 \\ \green{+4.05\%}}$ & $\substack{ 87.48 \\ \green{+2.64\%}}$ & $\substack{ 89.42 \\ \green{+0.77\%}}$ & $\substack{ 85.62 \\ \green{+4.43\%}}$ & $\substack{ 79.21 \\ \green{+2.01\%}}$ & $\substack{ 80.98 \\ \green{+0.95\%}}$ & $\substack{ 77.52 \\ \green{+3.03\%}}$ \\
        \midrule
        
        \multicolumn{10}{l}{\textit{Open-source Models}} \\
        AuroraCap-7B \citeyearpar{chai2024auroracap} & $\substack{ 27.00 \\ \red{-32.26\%}}$ & $\substack{ 57.24 \\ \red{-9.14\%}}$ & $\substack{ 17.67 \\ \red{-39.40\%}}$ & $\substack{ 29.31 \\ \red{-33.17\%}}$ & $\substack{ 62.35 \\ \red{-8.85\%}}$ & $\substack{ 19.16 \\ \red{-40.64\%}}$ & $\substack{ 21.15 \\ \red{-28.26\%}}$ & $\substack{ 44.46 \\ \red{-7.87\%}}$ & $\substack{ 13.88 \\ \red{-34.59\%}}$ \\
        LLaVA-Video-7B \citeyearpar{zhang2024videoinstructiontuningsynthetic} & $\substack{ 60.63 \\ \red{-4.01\%}}$ & $\substack{ 69.56 \\ \red{-5.50\%}}$ & $\substack{ 53.73 \\ \red{-2.84\%}}$ & $\substack{ 64.66 \\ \red{-3.25\%}}$ & $\substack{ 74.65 \\ \red{-4.32\%}}$ & $\substack{ 57.03 \\ \red{-2.43\%}}$ & $\substack{ 50.54 \\ \red{-6.15\%}}$ & $\substack{ 57.11 \\ \red{-8.64\%}}$ & $\substack{ 45.32 \\ \red{-4.19\%}}$ \\
        Tarsier2-7b \citeyearpar{yuan2025tarsier2advancinglargevisionlanguage} & $\substack{ 9.95 \\ \red{-76.52\%}}$ & $\substack{ 11.92 \\ \red{-78.64\%}}$ & $\substack{ 8.54 \\ \red{-75.00\%}}$ & $\substack{ 10.20 \\ \red{-76.58\%}}$ & $\substack{ 12.21 \\ \red{-78.92\%}}$ & $\substack{ 8.75 \\ \red{-74.93\%}}$ & $\substack{ 9.32 \\ \red{-76.37\%}}$ & $\substack{ 11.16 \\ \red{-77.99\%}}$ & $\substack{ 8.00 \\ \red{-75.21\%}}$ \\
        InternVL3.5-8B \citeyearpar{wang2025internvl3} & $\substack{ 64.03 \\ \green{+40.14\%}}$ & $\substack{ 71.89 \\ \green{+6.39\%}}$ & $\substack{ 57.72 \\ \green{+67.26\%}}$ & $\substack{ 67.53 \\ \green{+41.66\%}}$ & $\substack{ 76.05 \\ \green{+4.95\%}}$ & $\substack{ 60.73 \\ \green{+70.97\%}}$ & $\substack{ 55.20 \\ \green{+35.00\%}}$ & $\substack{ 61.51 \\ \green{+8.39\%}}$ & $\substack{ 50.06 \\ \green{+56.68\%}}$ \\
        InternVL3.5-30B-A3B \citeyearpar{wang2025internvl3} & $\substack{ 64.97 \\ \green{+19.30\%}}$ & $\substack{ 73.61 \\ \green{+5.11\%}}$ & $\substack{ 58.14 \\ \green{+30.51\%}}$ & $\substack{ 68.37 \\ \green{+18.04\%}}$ & $\substack{ 77.94 \\ \green{+3.88\%}}$ & $\substack{ 60.88 \\ \green{+29.09\%}}$ & $\substack{ 56.47 \\ \green{+23.30\%}}$ & $\substack{ 63.02 \\ \green{+8.97\%}}$ & $\substack{ 51.15 \\ \green{+34.89\%}}$ \\
        \midrule
        
        \multicolumn{10}{l}{\textit{Qwen3-VL Series} \citeyearpar{bai2025qwen3vltechnicalreport}} \\
        Qwen3-VL-2B & $\substack{ 70.46 \\ \green{+14.51\%}}$ & $\substack{ 75.99 \\ \green{+1.63\%}}$ & $\substack{ 65.67 \\ \green{+25.64\%}}$ & $\substack{ 74.31 \\ \green{+13.52\%}}$ & $\substack{ 80.10 \\ \green{+0.38\%}}$ & $\substack{ 69.30 \\ \green{+24.86\%}}$ & $\substack{ 60.64 \\ \green{+17.50\%}}$ & $\substack{ 65.49 \\ \green{+5.24\%}}$ & $\substack{ 56.46 \\ \green{+28.09\%}}$ \\
        Qwen3-VL-4B & $\substack{ 76.35 \\ \green{+8.84\%}}$ & $\substack{ 80.40 \\ \green{+2.64\%}}$ & $\substack{ 72.69 \\ \green{+14.45\%}}$ & $\substack{ 79.57 \\ \green{+8.33\%}}$ & $\substack{ 83.91 \\ \green{+2.06\%}}$ & $\substack{ 75.66 \\ \green{+13.98\%}}$ & $\substack{ 68.20 \\ \green{+10.37\%}}$ & $\substack{ 71.56 \\ \green{+4.33\%}}$ & $\substack{ 65.15 \\ \green{+15.88\%}}$ \\
        Qwen3-VL-8B & $\substack{ 78.22 \\ \green{+10.61\%}}$ & $\substack{ 81.82 \\ \green{+4.59\%}}$ & $\substack{ 74.92 \\ \green{+16.12\%}}$ & $\substack{ 81.03 \\ \green{+9.12\%}}$ & $\substack{ 84.87 \\ \green{+2.94\%}}$ & $\substack{ 77.52 \\ \green{+14.76\%}}$ & $\substack{ 71.10 \\ \green{+15.01\%}}$ & $\substack{ 74.13 \\ \green{+9.40\%}}$ & $\substack{ 68.30 \\ \green{+20.18\%}}$ \\
        Qwen3-VL-30B-A3B & $\substack{ 80.09 \\ \green{+16.38\%}}$ & $\substack{ 83.26 \\ \green{+6.33\%}}$ & $\substack{ 77.16 \\ \green{+25.67\%}}$ & $\substack{ 82.90 \\ \green{+15.15\%}}$ & $\substack{ 86.30 \\ \green{+4.71\%}}$ & $\substack{ 79.76 \\ \green{+24.82\%}}$ & $\substack{ 72.97 \\ \green{+19.78\%}}$ & $\substack{ 75.59 \\ \green{+10.82\%}}$ & $\substack{ 70.53 \\ \green{+28.14\%}}$ \\
        Qwen3-VL-32B & $\substack{ 81.09 \\ \green{+4.03\%}}$ & $\substack{ 83.57 \\ \green{+1.83\%}}$ & $\substack{ 78.75 \\ \green{+6.09\%}}$ & $\substack{ 83.43 \\ \green{+3.33\%}}$ & $\substack{ 86.00 \\ \green{+0.92\%}}$ & $\substack{ 81.00 \\ \green{+5.61\%}}$ & $\substack{ 75.15 \\ \green{+5.96\%}}$ & $\substack{ 77.40 \\ \green{+4.33\%}}$ & $\substack{ 73.04 \\ \green{+7.52\%}}$ \\

        \bottomrule
    \end{tabular}
    
    \caption{\textbf{Evaluation results under the detailed caption prompt setting.} The colored subscripts indicate the relative performance change compared to the baseline standardized prompt results reported in Table \ref{tab:main_results}. \textcolor{green!50!black}{Green} and \textcolor{red}{red} denote performance improvement and decline, respectively.}
    \label{tab:detail_results}
    \vspace{-4mm}
\end{table*}

\subsection{Evaluation on SOTA Models}
\label{sec:exp}

We evaluated several popular proprietary and open-source models.
To ensure a fair comparison, all models were queried with a standardized prompt: \texttt{``Describe this video in detail''}, with visual inputs uniformly sampled at 32 frames per video.
The quantitative results are summarized in Table \ref{tab:main_results}, revealing three key observations.

\textbf{Model Capabilities and Scaling Trend.}
{\benchmarkname} effectively differentiates model tiers.
Generally, proprietary models outperform open-source counterparts, with GPT-5.2 achieving the highest Overall {\benchmarkscore} score of $83.08$.
Within the open-source landscape, we observe a distinct scaling trend in the Qwen3-VL series.
As activated model size increases from 2B to 32B, performance improves generally (e.g., Overall {\benchmarkscore} rises from $61.53$ to $77.95$), underscoring that increased parameter count correlates strongly with video caption quality.

\textbf{The Trade-off between Factuality and Coverage.}
A pervasive trend across all models is that {\benchmarkprec} (\textit{Factuality}) consistently exceeds {\benchmarkrecall} (\textit{Coverage}).
For instance, GPT-4o achieves a high factuality of $78.13$ but a significantly lower coverage of $66.71$.
This suggests that current VLLMs exhibit a conservative generation strategy: \textit{they tend to generate trustworthy descriptions but often fail to exhaustively cover the salient visual details}.
This indicates room for improvement in increasing caption density without introducing hallucinations.

\textbf{The Reasoning Gap.}
Performance on Inferential questions consistently lags behind Descriptive ones, validating the hierarchical difficulty of our taxonomy.
Crucially, this performance gap widens significantly for smaller models.
While top-tier models like GPT-5.2 see a moderate degradation ($\sim$10\%) when transitioning from Descriptive to Inferential tasks, smaller models like AuroraCap-7B suffer a significant drop of $\sim$30\%.
This indicates that while visual recognition is becoming a baseline capability, complex visual reasoning remains the primary differentiator for superior VLLMs.

\subsection{Prompt Sensitivity Analysis}

It is widely recognized that VLLM performance can be heavily influenced by prompt design. 
To validate our findings, we experimented with a more complex and detailed caption prompt (see Appendix \ref{app:prompt}) in this section, comparing it against the concise standardized prompt from our main experiments. 
As shown in Table \ref{tab:detail_results}, these results reveal divergent sensitivities across models.
For instance, the Overall {\benchmarkscore} of the \texttt{Gemini-2.5-flash} decreased by $3.07\%$, while that of \texttt{GPT-4.1-2025-04-14} increased by $4.81\%$.
This phenomenon is also observed in open-source models.
While the \texttt{Qwen3-VL} series demonstrates consistent performance gains with the detailed prompt, other open-source models, such as \texttt{LLaVA-Video-7B}, exhibit a $4.01\%$ decrease, and \texttt{Tarsier2-7b} shows an even more substantial decline of $76.52\%$.

Crucially, however, the main conclusions of our benchmark remain robust against these variations. 
The key trends from the previous section (i.e., \textit{Model Capabilities and Scaling Trend}, \textit{The Trade-off between Factuality and Coverage}, and \textit{The Reasoning Gap}) persist regardless of the prompt complexity,  demonstrating that {\benchmarkname} effectively captures fundamental model capabilities independent of prompting strategies.

\section{Conclusion}

In this paper, we introduce {\benchmarkname}, a novel reference-free benchmark designed to assess video captioning quality via information fidelity.
By leveraging human-verified, fine-grained multiple-choice questions, it decouples and quantifies caption factuality and coverage.
Experiments demonstrate that {\benchmarkname} achieves robust alignment with human judgments compared to existing evaluators.
Moreover, our analysis exposes a critical limitation in current SOTA VLLMs: \textit{While exhibiting high factual precision, they often struggle with comprehensive coverage and show significant degradation on inferential tasks compared to descriptive ones.}
We envision {\benchmarkname} as a vital testbed to guide future research toward more robust and grounded video understanding models.


\section*{Limitations}

First, the question set of {\benchmarkname} may not be exhaustive.
Although we generate an average of {\avgqa} QA pairs per video to capture a wide range of visual information, guaranteeing the complete coverage of every visual detail in a complex video remains practically challenging.
Consequently, our coverage metric ({\benchmarkrecall}) serves as a proxy based on identified salient information rather than an absolute measure of total visual content.
In cases where videos contain extremely dense or subtle background details not captured by our QA generation pipeline, the reported coverage scores might be slightly overestimated.
Moreover, the current QA pairs are exclusively in English, which limits the evaluation of VLLMs in multilingual contexts.

Furthermore, to maintain a scalable and reference-free evaluation, we utilize \texttt{GPT-4.1} as the judge.
While exhibiting high alignment with human experts, the evaluation is inherently constrained by the judge model's upper bound and susceptible to the closed-source nature of the API, where model updates may affect reproducibility. 
Additionally, potential biases in the judge model regarding ambiguous visual descriptions could introduce noise. 

\section*{Ethical Considerations}

We prioritize ethical concerns associated with video content in benchmark construction, particularly regarding privacy and safety. 
To mitigate these risks, we curated video samples exclusively from established open-source datasets distributed under Creative Commons or compatible licenses, ensuring compliance with their usage policies.
Regarding the text annotation, while commercial APIs (e.g., GPT and Gemini) implement built-in safety guardrails, we acknowledge the residual risk of introducing model-inherent biases or toxicity. 
Furthermore, we conducted a rigorous manual inspection to filter out any content containing potential Not Safe For Work (NSFW) elements or sensitive Personally Identifiable Information (PII).
We ensured that all human annotators involved in this verification process were compensated at a rate exceeding the local minimum wage, adhering to fair labor practices.

\section*{Acknowledgments}

We would like to express our special thanks to Xiaoyi Ren for the exceptional support, insightful discussions, and detailed feedback throughout this project. 
We also thank Peng Zhang, Wencong Zhang, and Wentao Wu for their valuable discussions and careful reviews. 
We are grateful to Syed Khaja Naseeruddin Ahmed, Michael FitzMaurice, Gulsagar Jassar, Ryyan Mukarram, and the annotation team for their important efforts in data annotation and verification. 
We additionally thank Ying-Chang Cheng, Ganesh Nagarajan, Bruce Leng, and Snehal Nagmote for their outstanding infrastructure support for data processing and analysis.


\bibliography{custom}

\begin{thebibliography}{47}
\providecommand{\natexlab}[1]{#1}

\bibitem[{Anderson et~al.(2016)Anderson, Fernando, Johnson, and Gould}]{anderson2016spice}
Peter Anderson, Basura Fernando, Mark Johnson, and Stephen Gould. 2016.
\newblock Spice: Semantic propositional image caption evaluation.
\newblock In \emph{European conference on computer vision}, pages 382--398. Springer.

\bibitem[{Bai et~al.(2025)Bai, Cai, Chen, Chen, Chen, Cheng, Deng, Ding, Gao, Ge, Ge, Guo, Huang, Huang, Huang, Hui, Jiang, Li, Li, Li, Li, Lin, Lin, Liu, Liu, Liu, Liu, Liu, Liu, Lu, Luo, Lv, Men, Meng, Ren, Ren, Song, Sun, Tang, Tu, Wan, Wang, Wang, Wang, Wang, Xie, Xu, Xu, Xu, Yang, Yang, Yang, Yang, Yu, Zhang, Zhang, Zhang, Zheng, Zhong, Zhou, Zhou, Zhou, Zhu, and Zhu}]{bai2025qwen3vltechnicalreport}
Shuai Bai, Yuxuan Cai, Ruizhe Chen, Keqin Chen, Xionghui Chen, Zesen Cheng, Lianghao Deng, Wei Ding, Chang Gao, Chunjiang Ge, Wenbin Ge, Zhifang Guo, Qidong Huang, Jie Huang, Fei Huang, Binyuan Hui, Shutong Jiang, Zhaohai Li, Mingsheng Li, and 45 others. 2025.
\newblock \href {https://arxiv.org/abs/2511.21631} {Qwen3-vl technical report}.
\newblock \emph{Preprint}, arXiv:2511.21631.

\bibitem[{Banerjee and Lavie(2005)}]{banerjee2005meteor}
Satanjeev Banerjee and Alon Lavie. 2005.
\newblock Meteor: An automatic metric for mt evaluation with improved correlation with human judgments.
\newblock In \emph{Proceedings of the acl workshop on intrinsic and extrinsic evaluation measures for machine translation and/or summarization}, pages 65--72.

\bibitem[{Chai et~al.(2024)Chai, Song, Du, Meng, Madhavan, Bar-Tal, Hwang, Xie, and Manning}]{chai2024auroracap}
Wenhao Chai, Enxin Song, Yilun Du, Chenlin Meng, Vashisht Madhavan, Omer Bar-Tal, Jenq-Neng Hwang, Saining Xie, and Christopher~D Manning. 2024.
\newblock Auroracap: Efficient, performant video detailed captioning and a new benchmark.
\newblock \emph{arXiv preprint arXiv:2410.03051}.

\bibitem[{Chen and Dolan(2011)}]{chen2011collecting}
David Chen and William~B Dolan. 2011.
\newblock Collecting highly parallel data for paraphrase evaluation.
\newblock In \emph{Proceedings of the 49th annual meeting of the association for computational linguistics: human language technologies}, pages 190--200.

\bibitem[{Comanici et~al.(2025)Comanici, Bieber, Schaekermann, Pasupat, Sachdeva, Dhillon, Blistein, Ram, Zhang, Rosen et~al.}]{comanici2025gemini}
Gheorghe Comanici, Eric Bieber, Mike Schaekermann, Ice Pasupat, Noveen Sachdeva, Inderjit Dhillon, Marcel Blistein, Ori Ram, Dan Zhang, Evan Rosen, and 1 others. 2025.
\newblock Gemini 2.5: Pushing the frontier with advanced reasoning, multimodality, long context, and next generation agentic capabilities.
\newblock \emph{arXiv preprint arXiv:2507.06261}.

\bibitem[{Fang et~al.(2024)Fang, Mao, Duan, Zhao, Li, Lin, and Chen}]{fang2024mmbench}
Xinyu Fang, Kangrui Mao, Haodong Duan, Xiangyu Zhao, Yining Li, Dahua Lin, and Kai Chen. 2024.
\newblock Mmbench-video: A long-form multi-shot benchmark for holistic video understanding.
\newblock \emph{Advances in Neural Information Processing Systems}, 37:89098--89124.

\bibitem[{Fu et~al.(2025)Fu, Dai, Luo, Li, Ren, Zhang, Wang, Zhou, Shen, Zhang et~al.}]{fu2025video}
Chaoyou Fu, Yuhan Dai, Yongdong Luo, Lei Li, Shuhuai Ren, Renrui Zhang, Zihan Wang, Chenyu Zhou, Yunhang Shen, Mengdan Zhang, and 1 others. 2025.
\newblock Video-mme: The first-ever comprehensive evaluation benchmark of multi-modal llms in video analysis.
\newblock In \emph{Proceedings of the Computer Vision and Pattern Recognition Conference}, pages 24108--24118.

\bibitem[{Fu et~al.(2024)Fu, Ng, Jiang, and Liu}]{fu2024gptscore}
Jinlan Fu, See~Kiong Ng, Zhengbao Jiang, and Pengfei Liu. 2024.
\newblock Gptscore: Evaluate as you desire.
\newblock In \emph{Proceedings of the 2024 Conference of the North American Chapter of the Association for Computational Linguistics: Human Language Technologies (Volume 1: Long Papers)}, pages 6556--6576.

\bibitem[{Gwet(2001)}]{gwet2001handbook}
Kilem Gwet. 2001.
\newblock Handbook of inter-rater reliability: How to estimate the level of agreement between two or multiple raters.
\newblock \emph{Gaithersburg, MD: STATAXIS Publishing Company}.

\bibitem[{Hessel et~al.(2021)Hessel, Holtzman, Forbes, Bras, and Choi}]{hessel2021clipscore}
Jack Hessel, Ari Holtzman, Maxwell Forbes, Ronan~Le Bras, and Yejin Choi. 2021.
\newblock Clipscore: A reference-free evaluation metric for image captioning.
\newblock \emph{arXiv preprint arXiv:2104.08718}.

\bibitem[{Hu et~al.(2023)Hu, Chen, Zhang, and Jin}]{hu2023infometic}
Anwen Hu, Shizhe Chen, Liang Zhang, and Qin Jin. 2023.
\newblock Infometic: An informative metric for reference-free image caption evaluation.
\newblock \emph{arXiv preprint arXiv:2305.06002}.

\bibitem[{Hurst et~al.(2024)Hurst, Lerer, Goucher, Perelman, Ramesh, Clark, Ostrow, Welihinda, Hayes, Radford et~al.}]{hurst2024gpt}
Aaron Hurst, Adam Lerer, Adam~P Goucher, Adam Perelman, Aditya Ramesh, Aidan Clark, AJ~Ostrow, Akila Welihinda, Alan Hayes, Alec Radford, and 1 others. 2024.
\newblock Gpt-4o system card.
\newblock \emph{arXiv preprint arXiv:2410.21276}.

\bibitem[{Jiang et~al.(2019)Jiang, Huang, Zhang, Wang, Zhang, Gan, Diesner, and Gao}]{jiang2019tiger}
Ming Jiang, Qiuyuan Huang, Lei Zhang, Xin Wang, Pengchuan Zhang, Zhe Gan, Jana Diesner, and Jianfeng Gao. 2019.
\newblock Tiger: Text-to-image grounding for image caption evaluation.
\newblock \emph{arXiv preprint arXiv:1909.02050}.

\bibitem[{Jung and Kim(2025)}]{jung2025qeva}
Woojun Jung and Junyeong Kim. 2025.
\newblock Qeva: A reference-free evaluation metric for narrative video summarization with multimodal question answering.
\newblock In \emph{Findings of the Association for Computational Linguistics: EMNLP 2025}, pages 24632--24642.

\bibitem[{Kendall(1948)}]{kendall1948rank}
Maurice~George Kendall. 1948.
\newblock Rank correlation methods.

\bibitem[{Khattak et~al.(2025)Khattak, Naeem, Hassan, Naseer, Tombari, Khan, and Khan}]{khattak2025good}
Muhammad~Uzair Khattak, Muhammad~Ferjad Naeem, Jameel Hassan, Muzammal Naseer, Federico Tombari, Fahad~Shahbaz Khan, and Salman Khan. 2025.
\newblock How good is my video-lmm? complex video reasoning and robustness evaluation suite for video-lmms.
\newblock In \emph{Proceedings of the Computer Vision and Pattern Recognition Conference}, pages 3642--3651.

\bibitem[{Krishna et~al.(2017)Krishna, Hata, Ren, Fei-Fei, and Carlos~Niebles}]{krishna2017dense}
Ranjay Krishna, Kenji Hata, Frederic Ren, Li~Fei-Fei, and Juan Carlos~Niebles. 2017.
\newblock Dense-captioning events in videos.
\newblock In \emph{Proceedings of the IEEE international conference on computer vision}, pages 706--715.

\bibitem[{Lebron et~al.(2022)Lebron, Graham, McGuinness, Kouramas, and O’Connor}]{lebron2022bertha}
Luis Lebron, Yvette Graham, Kevin McGuinness, Konstantinos Kouramas, and Noel~E O’Connor. 2022.
\newblock Bertha: Video captioning evaluation via transfer-learned human assessment.
\newblock In \emph{Proceedings of the Thirteenth Language Resources and Evaluation Conference}, pages 1566--1575.

\bibitem[{Lee et~al.(2021)Lee, Scialom, Yoon, Dernoncourt, and Jung}]{lee2021qace}
Hwanhee Lee, Thomas Scialom, Seunghyun Yoon, Franck Dernoncourt, and Kyomin Jung. 2021.
\newblock Qace: Asking questions to evaluate an image caption.
\newblock \emph{arXiv preprint arXiv:2108.12560}.

\bibitem[{Lee et~al.(2018)Lee, Chen, Hua, Hu, and He}]{lee2018stacked}
Kuang-Huei Lee, Xi~Chen, Gang Hua, Houdong Hu, and Xiaodong He. 2018.
\newblock Stacked cross attention for image-text matching.
\newblock In \emph{Proceedings of the European conference on computer vision (ECCV)}, pages 201--216.

\bibitem[{Li et~al.(2024)Li, Wang, He, Li, Wang, Liu, Wang, Xu, Chen, Luo et~al.}]{li2024mvbench}
Kunchang Li, Yali Wang, Yinan He, Yizhuo Li, Yi~Wang, Yi~Liu, Zun Wang, Jilan Xu, Guo Chen, Ping Luo, and 1 others. 2024.
\newblock Mvbench: A comprehensive multi-modal video understanding benchmark.
\newblock In \emph{Proceedings of the IEEE/CVF Conference on Computer Vision and Pattern Recognition}, pages 22195--22206.

\bibitem[{Lin(2004)}]{lin2004rouge}
Chin-Yew Lin. 2004.
\newblock Rouge: A package for automatic evaluation of summaries.
\newblock In \emph{Text summarization branches out}, pages 74--81.

\bibitem[{Lin et~al.(2024)Lin, Pathak, Li, Li, Xia, Neubig, Zhang, and Ramanan}]{lin2024evaluating}
Zhiqiu Lin, Deepak Pathak, Baiqi Li, Jiayao Li, Xide Xia, Graham Neubig, Pengchuan Zhang, and Deva Ramanan. 2024.
\newblock Evaluating text-to-visual generation with image-to-text generation.
\newblock In \emph{European Conference on Computer Vision}, pages 366--384. Springer.

\bibitem[{Liu et~al.(2023)Liu, Iter, Xu, Wang, Xu, and Zhu}]{liu2023g}
Yang Liu, Dan Iter, Yichong Xu, Shuohang Wang, Ruochen Xu, and Chenguang Zhu. 2023.
\newblock G-eval: Nlg evaluation using gpt-4 with better human alignment.
\newblock \emph{arXiv preprint arXiv:2303.16634}.

\bibitem[{Maaz et~al.(2024)Maaz, Rasheed, Khan, and Khan}]{maaz2024video}
Muhammad Maaz, Hanoona Rasheed, Salman Khan, and Fahad Khan. 2024.
\newblock Video-chatgpt: Towards detailed video understanding via large vision and language models.
\newblock In \emph{Proceedings of the 62nd Annual Meeting of the Association for Computational Linguistics (Volume 1: Long Papers)}, pages 12585--12602.

\bibitem[{Miller(1995)}]{miller1995wordnet}
George~A Miller. 1995.
\newblock Wordnet: a lexical database for english.
\newblock \emph{Communications of the ACM}, 38(11):39--41.

\bibitem[{{OpenAI}(2025)}]{openai2025gpt51}
{OpenAI}. 2025.
\newblock Gpt-5.1: A smarter, more conversational chatgpt.
\newblock \url{https://openai.com/index/gpt-5-1/}.

\bibitem[{OpenAI(2025)}]{oai2025gpt41}
OpenAI. 2025.
\newblock Introducing gpt-4.1 in the api.
\newblock \url{https://openai.com/index/gpt-4-1/}.
\newblock Accessed: 2025-04-14.

\bibitem[{Papineni et~al.(2002)Papineni, Roukos, Ward, and Zhu}]{papineni2002bleu}
Kishore Papineni, Salim Roukos, Todd Ward, and Wei-Jing Zhu. 2002.
\newblock Bleu: a method for automatic evaluation of machine translation.
\newblock In \emph{Proceedings of the 40th annual meeting of the Association for Computational Linguistics}, pages 311--318.

\bibitem[{Patraucean et~al.(2023)Patraucean, Smaira, Gupta, Recasens, Markeeva, Banarse, Koppula, Malinowski, Yang, Doersch et~al.}]{patraucean2023perception}
Viorica Patraucean, Lucas Smaira, Ankush Gupta, Adria Recasens, Larisa Markeeva, Dylan Banarse, Skanda Koppula, Mateusz Malinowski, Yi~Yang, Carl Doersch, and 1 others. 2023.
\newblock Perception test: A diagnostic benchmark for multimodal video models.
\newblock \emph{Advances in Neural Information Processing Systems}, 36:42748--42761.

\bibitem[{Radford et~al.(2021)Radford, Kim, Hallacy, Ramesh, Goh, Agarwal, Sastry, Askell, Mishkin, Clark et~al.}]{radford2021learning}
Alec Radford, Jong~Wook Kim, Chris Hallacy, Aditya Ramesh, Gabriel Goh, Sandhini Agarwal, Girish Sastry, Amanda Askell, Pamela Mishkin, Jack Clark, and 1 others. 2021.
\newblock Learning transferable visual models from natural language supervision.
\newblock In \emph{International conference on machine learning}, pages 8748--8763. PmLR.

\bibitem[{Spearman(1961)}]{spearman1961general}
Charles Spearman. 1961.
\newblock " general intelligence" objectively determined and measured.

\bibitem[{Vedantam et~al.(2015)Vedantam, Lawrence~Zitnick, and Parikh}]{vedantam2015cider}
Ramakrishna Vedantam, C~Lawrence~Zitnick, and Devi Parikh. 2015.
\newblock Cider: Consensus-based image description evaluation.
\newblock In \emph{Proceedings of the IEEE conference on computer vision and pattern recognition}, pages 4566--4575.

\bibitem[{Wang et~al.(2024)Wang, Yuan, Zhang, and Sun}]{wang2024tarsier}
Jiawei Wang, Liping Yuan, Yuchen Zhang, and Haomiao Sun. 2024.
\newblock Tarsier: Recipes for training and evaluating large video description models.
\newblock \emph{arXiv preprint arXiv:2407.00634}.

\bibitem[{Wang et~al.(2020)Wang, Wang, Yao, Shan, and Chen}]{wang2020cross}
Sijin Wang, Ruiping Wang, Ziwei Yao, Shiguang Shan, and Xilin Chen. 2020.
\newblock Cross-modal scene graph matching for relationship-aware image-text retrieval.
\newblock In \emph{Proceedings of the IEEE/CVF winter conference on applications of computer vision}, pages 1508--1517.

\bibitem[{Wang et~al.(2021)Wang, Yao, Wang, Wu, and Chen}]{wang2021faier}
Sijin Wang, Ziwei Yao, Ruiping Wang, Zhongqin Wu, and Xilin Chen. 2021.
\newblock Faier: Fidelity and adequacy ensured image caption evaluation.
\newblock In \emph{Proceedings of the IEEE/CVF Conference on Computer Vision and Pattern Recognition}, pages 14050--14059.

\bibitem[{Wang et~al.(2025)Wang, Gao, Gu, Pu, Cui, Wei, Liu, Jing, Ye, Shao et~al.}]{wang2025internvl3}
Weiyun Wang, Zhangwei Gao, Lixin Gu, Hengjun Pu, Long Cui, Xingguang Wei, Zhaoyang Liu, Linglin Jing, Shenglong Ye, Jie Shao, and 1 others. 2025.
\newblock Internvl3. 5: Advancing open-source multimodal models in versatility, reasoning, and efficiency.
\newblock \emph{arXiv preprint arXiv:2508.18265}.

\bibitem[{Wang et~al.(2019)Wang, Wu, Chen, Li, Wang, and Wang}]{wang2019vatex}
Xin Wang, Jiawei Wu, Junkun Chen, Lei Li, Yuan-Fang Wang, and William~Yang Wang. 2019.
\newblock Vatex: A large-scale, high-quality multilingual dataset for video-and-language research.
\newblock In \emph{Proceedings of the IEEE/CVF international conference on computer vision}, pages 4581--4591.

\bibitem[{Wu et~al.(2024)Wu, Li, Chen, and Li}]{wu2024longvideobench}
Haoning Wu, Dongxu Li, Bei Chen, and Junnan Li. 2024.
\newblock Longvideobench: A benchmark for long-context interleaved video-language understanding.
\newblock \emph{Advances in Neural Information Processing Systems}, 37:28828--28857.

\bibitem[{Xiao et~al.(2021)Xiao, Shang, Yao, and Chua}]{xiao2021next}
Junbin Xiao, Xindi Shang, Angela Yao, and Tat-Seng Chua. 2021.
\newblock Next-qa: Next phase of question-answering to explaining temporal actions.
\newblock In \emph{Proceedings of the IEEE/CVF conference on computer vision and pattern recognition}, pages 9777--9786.

\bibitem[{Xu et~al.(2016)Xu, Mei, Yao, and Rui}]{xu2016msr}
Jun Xu, Tao Mei, Ting Yao, and Yong Rui. 2016.
\newblock Msr-vtt: A large video description dataset for bridging video and language.
\newblock In \emph{Proceedings of the IEEE conference on computer vision and pattern recognition}, pages 5288--5296.

\bibitem[{Yang et~al.(2025)Yang, Liu, Zhai, Sun, Liu, Barsoum, Li, and Xu}]{yang2025captionqa}
Shijia Yang, Yunong Liu, Bohan Zhai, Ximeng Sun, Zicheng Liu, Emad Barsoum, Manling Li, and Chenfeng Xu. 2025.
\newblock Captionqa: Is your caption as useful as the image itself?
\newblock \emph{arXiv preprint arXiv:2511.21025}.

\bibitem[{Yuan et~al.(2025)Yuan, Wang, Sun, Zhang, and Lin}]{yuan2025tarsier2advancinglargevisionlanguage}
Liping Yuan, Jiawei Wang, Haomiao Sun, Yuchen Zhang, and Yuan Lin. 2025.
\newblock \href {https://arxiv.org/abs/2501.07888} {Tarsier2: Advancing large vision-language models from detailed video description to comprehensive video understanding}.
\newblock \emph{Preprint}, arXiv:2501.07888.

\bibitem[{Zhang et~al.(2025)Zhang, Wang, Huang, Li, Zhu, and Yin}]{zhang2025vcapsbench}
Shi-Xue Zhang, Hongfa Wang, Duojun Huang, Xin Li, Xiaobin Zhu, and Xu-Cheng Yin. 2025.
\newblock Vcapsbench: A large-scale fine-grained benchmark for video caption quality evaluation.
\newblock \emph{arXiv preprint arXiv:2505.23484}.

\bibitem[{Zhang et~al.(2019)Zhang, Kishore, Wu, Weinberger, and Artzi}]{zhang2019bertscore}
Tianyi Zhang, Varsha Kishore, Felix Wu, Kilian~Q Weinberger, and Yoav Artzi. 2019.
\newblock Bertscore: Evaluating text generation with bert.
\newblock \emph{arXiv preprint arXiv:1904.09675}.

\bibitem[{Zhang et~al.(2024)Zhang, Wu, Li, Li, Ma, Liu, and Li}]{zhang2024videoinstructiontuningsynthetic}
Yuanhan Zhang, Jinming Wu, Wei Li, Bo~Li, Zejun Ma, Ziwei Liu, and Chunyuan Li. 2024.
\newblock \href {https://arxiv.org/abs/2410.02713} {Video instruction tuning with synthetic data}.
\newblock \emph{Preprint}, arXiv:2410.02713.

\end{thebibliography}

\appendix

\section{Robustness to Judge Choice}

\begin{table*}[t]
    \centering
    \small 
    \setlength{\tabcolsep}{4.5pt} 
    \renewcommand{\arraystretch}{1.5}
    
    \begin{tabular}{l ccc ccc ccc}
        \toprule
        \multirow{2}{*}{\textbf{Model}} & \multicolumn{3}{c}{\textbf{Overall}} & \multicolumn{3}{c}{\textbf{Descriptive}} & \multicolumn{3}{c}{\textbf{Inferential}} \\
        \cmidrule(lr){2-4} \cmidrule(lr){5-7} \cmidrule(lr){8-10}
         & F1 & P & R & F1 & P & R & F1 & P & R \\
        \midrule

        \multicolumn{10}{l}{\textit{Proprietary Models}} \\
        Gemini-2.5-flash & $\substack{ 74.84 \\ \red{-7.62\%}}$ & $\substack{ 82.38 \\ \red{-2.53\%}}$ & $\substack{ 68.57 \\ \red{-11.85\%}}$ & $\substack{ 78.47 \\ \red{-6.42\%}}$ & $\substack{ 86.76 \\ \red{-1.06\%}}$ & $\substack{ 71.63 \\ \red{-10.84\%}}$ & $\substack{ 65.74 \\ \red{-10.97\%}}$ & $\substack{ 71.56 \\ \red{-6.56\%}}$ & $\substack{ 60.79 \\ \red{-14.73\%}}$ \\
        Gemini-2.5-pro & $\substack{ 77.43 \\ \red{-5.24\%}}$ & $\substack{ 83.43 \\ \red{-1.78\%}}$ & $\substack{ 72.22 \\ \red{-8.26\%}}$ & $\substack{ 80.84 \\ \red{-4.10\%}}$ & $\substack{ 87.14 \\ \red{-0.68\%}}$ & $\substack{ 75.40 \\ \red{-7.04\%}}$ & $\substack{ 68.75 \\ \red{-8.53\%}}$ & $\substack{ 74.03 \\ \red{-4.94\%}}$ & $\substack{ 64.18 \\ \red{-11.63\%}}$ \\
        GPT-4o-2024-11-20 & $\substack{ 66.76 \\ \red{-7.24\%}}$ & $\substack{ 79.30 \\ \green{+1.50\%}}$ & $\substack{ 57.64 \\ \red{-13.60\%}}$ & $\substack{ 70.59 \\ \red{-5.87\%}}$ & $\substack{ 84.04 \\ \green{+2.45\%}}$ & $\substack{ 60.85 \\ \red{-11.90\%}}$ & $\substack{ 57.08 \\ \red{-11.44\%}}$ & $\substack{ 67.43 \\ \red{-1.85\%}}$ & $\substack{ 49.49 \\ \red{-18.47\%}}$ \\
        GPT-4.1-2025-04-14 & $\substack{ 71.83 \\ \red{-6.46\%}}$ & $\substack{ 80.76 \\ \red{-1.09\%}}$ & $\substack{ 64.67 \\ \red{-10.76\%}}$ & $\substack{ 75.44 \\ \red{-4.92\%}}$ & $\substack{ 85.30 \\ \green{+0.57\%}}$ & $\substack{ 67.62 \\ \red{-9.27\%}}$ & $\substack{ 62.78 \\ \red{-10.82\%}}$ & $\substack{ 69.62 \\ \red{-5.77\%}}$ & $\substack{ 57.16 \\ \red{-14.99\%}}$ \\
        GPT-5.2-2025-12-11 & $\substack{ 78.49 \\ \red{-5.52\%}}$ & $\substack{ 84.45 \\ \red{-2.16\%}}$ & $\substack{ 73.31 \\ \red{-8.47\%}}$ & $\substack{ 81.60 \\ \red{-4.26\%}}$ & $\substack{ 87.95 \\ \red{-0.89\%}}$ & $\substack{ 76.10 \\ \red{-7.18\%}}$ & $\substack{ 70.62 \\ \red{-9.05\%}}$ & $\substack{ 75.64 \\ \red{-5.71\%}}$ & $\substack{ 66.23 \\ \red{-11.98\%}}$ \\
        \midrule
        
        \multicolumn{10}{l}{\textit{Open-source Models}} \\
        AuroraCap-7B & $\substack{ 33.91 \\ \red{-14.93\%}}$ & $\substack{ 64.47 \\ \green{+2.33\%}}$ & $\substack{ 23.01 \\ \red{-21.09\%}}$ & $\substack{ 38.05 \\ \red{-13.25\%}}$ & $\substack{ 68.73 \\ \green{+0.48\%}}$ & $\substack{ 26.31 \\ \red{-18.49\%}}$ & $\substack{ 22.37 \\ \red{-24.12\%}}$ & $\substack{ 49.81 \\ \green{+3.21\%}}$ & $\substack{ 14.42 \\ \red{-32.05\%}}$ \\
        LLaVA-Video-7B & $\substack{ 56.57 \\ \red{-10.43\%}}$ & $\substack{ 74.15 \\ \green{+0.73\%}}$ & $\substack{ 45.73 \\ \red{-17.31\%}}$ & $\substack{ 61.09 \\ \red{-8.59\%}}$ & $\substack{ 79.07 \\ \green{+1.35\%}}$ & $\substack{ 49.77 \\ \red{-14.85\%}}$ & $\substack{ 44.46 \\ \red{-17.44\%}}$ & $\substack{ 60.33 \\ \red{-3.49\%}}$ & $\substack{ 35.20 \\ \red{-25.58\%}}$ \\
        Tarsier2-7b & $\substack{ 43.82 \\ \green{+3.42\%}}$ & $\substack{ 68.77 \\ \green{+23.24\%}}$ & $\substack{ 32.16 \\ \red{-5.85\%}}$ & $\substack{ 45.61 \\ \green{+4.73\%}}$ & $\substack{ 74.39 \\ \green{+28.46\%}}$ & $\substack{ 32.89 \\ \red{-5.76\%}}$ & $\substack{ 39.47 \\ \green{+0.08\%}}$ & $\substack{ 56.71 \\ \green{+11.85\%}}$ & $\substack{ 30.27 \\ \red{-6.20\%}}$ \\
        InternVL3.5-8B & $\substack{ 39.66 \\ \red{-13.20\%}}$ & $\substack{ 70.27 \\ \green{+4.00\%}}$ & $\substack{ 27.63 \\ \red{-19.94\%}}$ & $\substack{ 42.36 \\ \red{-11.14\%}}$ & $\substack{ 74.54 \\ \green{+2.87\%}}$ & $\substack{ 29.59 \\ \red{-16.69\%}}$ & $\substack{ 32.63 \\ \red{-20.20\%}}$ & $\substack{ 58.84 \\ \green{+3.68\%}}$ & $\substack{ 22.57 \\ \red{-29.36\%}}$ \\
        InternVL3.5-30B-A3B & $\substack{ 48.18 \\ \red{-11.53\%}}$ & $\substack{ 72.28 \\ \green{+3.21\%}}$ & $\substack{ 36.14 \\ \red{-18.88\%}}$ & $\substack{ 52.03 \\ \red{-10.17\%}}$ & $\substack{ 76.68 \\ \green{+2.20\%}}$ & $\substack{ 39.38 \\ \red{-16.50\%}}$ & $\substack{ 37.83 \\ \red{-17.40\%}}$ & $\substack{ 59.61 \\ \green{+3.08\%}}$ & $\substack{ 27.71 \\ \red{-26.93\%}}$ \\
        \midrule
        
        \multicolumn{10}{l}{\textit{Qwen3-VL Series} \citeyearpar{bai2025qwen3vltechnicalreport}} \\
        Qwen3-VL-2B & $\substack{ 56.56 \\ \red{-8.08\%}}$ & $\substack{ 74.79 \\ \green{+0.03\%}}$ & $\substack{ 45.48 \\ \red{-12.99\%}}$ & $\substack{ 60.78 \\ \red{-7.15\%}}$ & $\substack{ 79.75 \\ \red{-0.06\%}}$ & $\substack{ 49.10 \\ \red{-11.53\%}}$ & $\substack{ 45.65 \\ \red{-11.55\%}}$ & $\substack{ 61.60 \\ \red{-1.01\%}}$ & $\substack{ 36.26 \\ \red{-17.74\%}}$ \\
        Qwen3-VL-4B & $\substack{ 64.69 \\ \red{-7.78\%}}$ & $\substack{ 77.94 \\ \red{-0.50\%}}$ & $\substack{ 55.29 \\ \red{-12.94\%}}$ & $\substack{ 68.75 \\ \red{-6.40\%}}$ & $\substack{ 82.72 \\ \green{+0.61\%}}$ & $\substack{ 58.82 \\ \red{-11.39\%}}$ & $\substack{ 54.31 \\ \red{-12.11\%}}$ & $\substack{ 65.68 \\ \red{-4.24\%}}$ & $\substack{ 46.30 \\ \red{-17.64\%}}$ \\
        Qwen3-VL-8B & $\substack{ 66.04 \\ \red{-6.62\%}}$ & $\substack{ 78.46 \\ \green{+0.29\%}}$ & $\substack{ 57.02 \\ \red{-11.62\%}}$ & $\substack{ 70.08 \\ \red{-5.63\%}}$ & $\substack{ 83.14 \\ \green{+0.84\%}}$ & $\substack{ 60.56 \\ \red{-10.35\%}}$ & $\substack{ 55.76 \\ \red{-9.80\%}}$ & $\substack{ 66.45 \\ \red{-1.93\%}}$ & $\substack{ 48.04 \\ \red{-15.47\%}}$ \\
        Qwen3-VL-30B-A3B & $\substack{ 63.15 \\ \red{-8.24\%}}$ & $\substack{ 77.36 \\ \red{-1.20\%}}$ & $\substack{ 53.36 \\ \red{-13.09\%}}$ & $\substack{ 66.86 \\ \red{-7.13\%}}$ & $\substack{ 81.90 \\ \red{-0.63\%}}$ & $\substack{ 56.49 \\ \red{-11.60\%}}$ & $\substack{ 53.72 \\ \red{-11.82\%}}$ & $\substack{ 65.80 \\ \red{-3.53\%}}$ & $\substack{ 45.38 \\ \red{-17.55\%}}$ \\
        Qwen3-VL-32B & $\substack{ 73.36 \\ \red{-5.89\%}}$ & $\substack{ 80.84 \\ \red{-1.50\%}}$ & $\substack{ 67.14 \\ \red{-9.55\%}}$ & $\substack{ 77.08 \\ \red{-4.53\%}}$ & $\substack{ 84.96 \\ \red{-0.31\%}}$ & $\substack{ 70.54 \\ \red{-8.03\%}}$ & $\substack{ 63.90 \\ \red{-9.90\%}}$ & $\substack{ 70.38 \\ \red{-5.14\%}}$ & $\substack{ 58.51 \\ \red{-13.87\%}}$ \\

        \bottomrule
    \end{tabular}
    
    \caption{\textbf{Evaluation results with \texttt{Qwen3-30B-A3B-Instruct-2507} as the Judge.} The colored subscripts indicate relative changes with respect to the \texttt{GPT-4.1}-based results in Table \ref{tab:main_results}. \textcolor{green!50!black}{Green} and \textcolor{red}{red} denote performance improvement and decline, respectively.}
    \label{tab:qwen_results}
    \vspace{-4mm}
\end{table*}

To ensure reproducibility, all main experiments use a fixed API snapshot of \texttt{GPT-4.1} (i.e. gpt-4.1-2025-04-14) with deterministic decoding.
While \texttt{GPT-4.1} serves as the default automatic judge due to its strong instruction-following and stable answer formatting, CapQuiz itself is not tied to a specific judge model. 
To examine the sensitivity of CapQuiz to judge choice, we additionally instantiate the same evaluation pipeline with a strong open-weight model, \texttt{Qwen3-30B-A3B-Instruct-2507}.
We keep the prompt template, answer space, and parsing rules identical across judges, and re-evaluate the VLLMs in Section \ref{sec:exp} under the Qwen3-based judge.

As summarized in Table \ref{tab:qwen_results}, we observe two main findings. 
First, replacing \texttt{GPT-4.1} with \texttt{Qwen3-30B-A3B-Instruct-2507} preserves the overall ranking trends of VLLMs and leaves the main conclusions of Section \ref{sec:exp} unchanged.
This suggests that the comparative findings of CapQuiz are not tied to a single proprietary judge. 
Second, we observe a general decrease in absolute CapP, CapR, and CapF1 scores under \texttt{Qwen3-30B-A3B-Instruct-2507}. 
This shift likely reflects differences in instruction-following and zero-shot reasoning capability across judges.
Overall, these results suggest that GPT-4.1 is a strong default judge that improves score calibration, while the main comparative conclusions of CapQuiz remain robust to judge substitution.

\section{Robustness to QA Set Size}

CapR is intended as a proxy for the coverage of salient visual information, rather than an exhaustive recall measure over all possible video details. 
Since each video is annotated with a finite set of curated QA pairs, it is important to verify that the resulting evaluation remains stable with respect to the number of questions used. 
We therefore conduct an ablation study by varying the retained proportion of QA pairs per video.

Specifically, for each video, we uniformly subsample its QA set with retention ratios in {25\%, 50\%, 75\%, 100\%}.
For the 25\%, 50\%, and 75\% settings, we repeat the subsampling with 10 random seeds and report the mean and standard deviation of CapF1.
We evaluate four representative VLLMs spanning different model families and capability levels: \texttt{Gemini-2.5-pro}, \texttt{GPT-5.2}, \texttt{Qwen3-VL-30B-A3B}, and \texttt{Qwen3-VL-32B}.

As shown in Table \ref{tab:qa_retention_ablation}, we observe two main findings. 
First, the comparative conclusions are highly stable across all retention ratios: even under the aggressive 25\% setting, the relative ranking of the evaluated VLLMs remains unchanged. 
This indicates that CapQuiz is not overly sensitive to the exact number of QA pairs used for evaluation at the system-comparison level. 
Second, the score variance exhibits a clear threshold effect. 
At 50\% and 75\% retention, the standard deviation of CapF1 remains very small across all evaluated models, whereas the 25\% setting leads to noticeably larger variance. 
These results suggest that the default annotation density in CapQuiz (19+ QA pairs per video on average) is safely above the stability threshold needed for reliable evaluation.
Overall, this analysis supports the interpretation of CapR as a salience-oriented coverage proxy: although it is not exhaustive by design, the current QA density is sufficient to yield stable model comparisons and robust CapF1 estimates.

\begin{table}[t]
    \centering
    \small
    \resizebox{\columnwidth}{!}{
        \begin{tabular}{lcccc}
            \toprule
            \textbf{Retention Rate} & \texttt{gemini-2.5-pro} & \texttt{gpt-5.2} & \texttt{Qwen3-VL-30B-A3B} & \texttt{Qwen3-VL-32B} \\
            \midrule
            25\%  & 81.72 $\pm$ 0.16 & 83.00 $\pm$ 0.07 & 68.90 $\pm$ 0.19 & 77.94 $\pm$ 0.21 \\
            50\%  & 81.78 $\pm$ 0.07 & 83.03 $\pm$ 0.03 & 68.57 $\pm$ 0.09 & 78.00 $\pm$ 0.08 \\
            75\%  & 81.64 $\pm$ 0.03 & 83.06 $\pm$ 0.01 & 68.80 $\pm$ 0.03 & 77.93 $\pm$ 0.03 \\
            100\%  & 81.71 & 83.08 & 68.82 & 77.95 \\
            \bottomrule
        \end{tabular}
    }
    \caption{\textbf{CapF1 under different QA retention ratios.} For each video, we uniformly subsample 25\%, 50\%, or 75\% of its QA pairs and repeat the evaluation with 10 random seeds. Results are reported as mean $\pm$ standard deviation. The 100\% row uses the full QA set.}
    \label{tab:qa_retention_ablation}
\end{table}

\section{Taxonomy}
\label{app:taxonomy}

\subsection{Video Taxonomy}
\label{app:vidtax}

\textbf{Knowledge} Content that systematically presents knowledge about the natural world, human societies, historical developments, and scientific or technological principles. The primary purpose is to inform, explain, and deepen understanding through structured, evidence-based narratives.
    \begin{itemize}[leftmargin=1em]
        \item \textbf{Nature}: Content about the natural world on Earth, including ecosystems, wildlife, environmental processes, and conservation efforts. Focuses on non-human-driven phenomena and the interdependence of living organisms and their habitats.
        \item \textbf{Science}: Systematic knowledge of the physical and technological world, including fundamental principles in physics, chemistry, biology, astronomy, and engineering. Covers how things work, from subatomic particles to space exploration, and the development of technologies such as AI and robotics.
        \item \textbf{Health}: Knowledge about the human body, medical science, disease prevention, mental well-being, and public health. Emphasizes evidence-based understanding of health conditions, treatments, and lifestyle impacts on physical and psychological wellness.
        \item \textbf{History}: Documented understanding of past human events, civilizations, conflicts, discoveries, and cultural developments. Based on historical records, archaeological findings, and scholarly analysis of how societies have evolved over time.
        \item \textbf{Society}: Insights into human social structures, behaviors, institutions, and collective thought. Includes economics, psychology, education, ethics, philosophy, and the study of how individuals and groups interact within cultural and organizational contexts.
    \end{itemize}
\textbf{Everyday} Authentic recordings of ordinary life that capture personal experiences, relationships with people and animals, and moments of solitude. Focuses on unscripted, non-performance-based content that reflects how individuals live, connect, and exist in their daily environments—whether alone, with others, or alongside companion animals.
    \begin{itemize}[leftmargin=1em]
        \item \textbf{Human Bonds}: Authentic moments of connection and coexistence with family, friends, partners, or acquaintances, emphasizing emotional intimacy, shared experiences, and the warmth of human relationships.
        \item \textbf{Animal Companions}: Daily life and emotional bonding between humans and their animal companions, highlighting care, spontaneity, and the unique non-verbal intimacy shared across species.
        \item \textbf{Personal Life}: Recordings of an individual’s daily existence in solitude, encompassing routines, habits, reflections, emotions, domestic activities, and atmospheric moments. Focuses on how a person experiences, manages, and expresses their life without interaction with people or pets. This includes personal journeys, functional tasks, and contemplative states, all centered on the self as the sole subject.
    \end{itemize}
\textbf{Creativity} Content that expresses imagination, emotion, or aesthetic vision through artistic performance, storytelling, or physical excellence. Includes movies, music, dance, animation, comedy, and sports events. The primary intent is to be seen, heard, or experienced as a form of personal or collaborative expression—not for instruction, commerce, or information alone.
    \begin{itemize}[leftmargin=1em]
        \item \textbf{Movie \& Show}: Fictional or dramatic videos that tell a story, including movies, TV series, web dramas, and short films. Typically feature actors, scripts, and narrative structure.
        \item \textbf{Dance \& Performance}: Choreographed or expressive performances centered on movement, including original dance routines, dance covers, stage shows, spoken word poetry, and artistic recitations. Emphasizes physical expression, rhythm, and emotional delivery.
        \item \textbf{Music \& Singing}: Original or performed musical works, including official music videos, song releases, vocal covers, instrumental performances, and creative audio-visual compositions. Focuses on auditory artistry and musical expression.
        \item \textbf{Animation}: Animated works created through 2D, 3D, stop-motion, or digital techniques, including short films, creative explainers, and experimental visual stories. Emphasizes visual imagination and motion design.
        \item \textbf{Comedy Sketch}: Short, scripted humorous videos designed to entertain, including parodies, satirical scenes, original comedy skits, and creative spoofs. Often feature exaggerated characters and comedic timing.
        \item \textbf{Game}: Creative content made within or about games, such as custom maps, mods, character designs, in-game art projects, or narrative-driven gameplay. Emphasizes originality, design, and virtual world-building.
        \item \textbf{Sports}: Content centered on athletic competitions and physical performance, including live events, highlights, athlete stories, news, and expert analysis. Emphasizes the drama, skill, and emotional intensity of sports as a form of visual and emotional entertainment.
    \end{itemize}
\textbf{Civics} Coverage of real-world public events, societal issues, political developments, and collective experiences that impact communities or nations. Focuses on factual reporting, public discourse, and awareness of civic life.
    \begin{itemize}[leftmargin=1em]
        \item \textbf{News}: Reporting on recent, impactful public events such as natural disasters, accidents, conflicts, or major societal incidents. Focuses on what happened, where, and when, with emphasis on timeliness and factual accuracy.
        \item \textbf{Social Issues}: Coverage of ongoing societal challenges and public debates, such as education inequality, mental health awareness, housing affordability, gender rights, racial justice, and environmental policy. Focuses on current events, stakeholder perspectives, and civic discourse.
        \item \textbf{Civic Action}: Recordings of collective efforts to address social or environmental issues, such as protests, volunteer work, humanitarian aid, and community organizing. Highlights public participation and social change.
    \end{itemize}
\textbf{Function} Content designed to help users accomplish a practical goal, such as learning how to cook a meal, perform a task, make a purchase decision, plan a trip, organize daily life, or review a recording for reference. The primary intent is utility—providing actionable guidance, decision support, or functional documentation—rather than entertainment, knowledge explanation, or personal expression.
    \begin{itemize}[leftmargin=1em]
        \item \textbf{How-To}: Step-by-step instructions for completing practical tasks in daily life, work, or learning—excluding cooking—such as repairing a device, using software, organizing space, crafting, or performing a physical skill. Covers both short-term actions and repeatable routines, with a focus on actionable guidance and immediate application.
        \item \textbf{Cooking}: Step-by-step instructions for preparing meals, dishes, or beverages, including recipe demonstrations, cooking techniques, meal prep, and kitchen tips. Focuses on food creation, flavor development, and practical kitchen skills.
        \item \textbf{Buyer's Guide}: Content that helps viewers decide what to buy, including product reviews, comparisons, recommendations, unboxing, and live commerce. Emphasizes real-world usage, value assessment, and decision support.
        \item \textbf{Travel Planning}: Guides for designing a trip, including itinerary creation, budgeting, transportation, accommodation, and visa planning. Helps viewers prepare for travel with practical, organized advice.
        \item \textbf{Life Guide}: Guides that help viewers design sustainable, personalized life systems, such as minimalism, daily routines, or personal workflows. Focuses on the philosophy, structure, and long-term optimization of everyday living—beyond step-by-step instructions.
        \item \textbf{Functional Recordings}: Videos recorded for practical purposes, such as screen recordings, surveillance, dashcams, meeting logs, or training replays. Not intended for entertainment or artistic expression.
    \end{itemize}

\subsection{Question Taxonomy}
\label{app:questax}

\textbf{Descriptive} Focus on factual information that is directly observable in the video content.
    \begin{itemize}[leftmargin=1em]
        \item \textbf{Entity}: Pertains to the identification of specific people, animals, objects, texts, or symbols present in the scene.
        \item \textbf{Attribute}: Involves visual properties (e.g., color, shape), quantity counts, or the physical states (e.g., open/closed) of the entities.
        \item \textbf{Action}: Captures physical movements, simple behaviors performed by a single entity, or direct physical interactions between entities.
        \item \textbf{Event}: Summarizes the composite activity, situation, or overarching happening depicted throughout the video clip.
        \item \textbf{Setting}: Describes the environmental context, background scenery, or visual cues indicating location and time of day.
    \end{itemize}
\textbf{Inferential} Focus on abstract information derived from visual cues through logical reasoning or contextual understanding.
    \begin{itemize}[leftmargin=1em]
        \item \textbf{Relation}: Captures the spatial configurations, temporal ordering, or comparative relationships between entities or actions.
        \item \textbf{Causality}: Explains the logic behind events, including causes (why?), immediate effects (results), or potential counterfactuals.
        \item \textbf{Intent \& Emotion}: Implies the agents' underlying goals, motivations, emotional states, or the psychological purpose behind their actions.
        \item \textbf{Thematic \& Symbolic}: Involves high-level understanding of the plot, abstract themes, or symbolic meanings embedded in the content.
        \item \textbf{Spatiotemporal Reference}: Includes specific deictic cues or references (e.g., "on the left", "at the beginning") that ground the text to precise spatial regions or temporal segments.
    \end{itemize}

\section{Prompts}
\label{app:prompt}

\begin{itemize}[leftmargin=1em]
    \item Video Taxonomy Prompt, see Figure \ref{fig:vid_tax_prompt}
    \item QA Generation Prompt, see Figure \ref{fig:qa_gen_prompt}
    \item Blind Solvability Check Prompt, see Figure \ref{fig:to_dedep}
    \item Question Deduplication Prompt, see Figure \ref{fig:to_dedep}
    \item Multiple-Choice QA Prompt, see Figure \ref{fig:grade_prompt}
    \item VLLM-as-a-Judge Prompt, see Figure \ref{fig:grade_prompt}
    \item Detailed Caption Prompt, see Figure \ref{fig:detail_prompt}
\end{itemize}

\begin{figure*}[b] 
    \centering
    \begin{tcolorbox}[
        title=\textbf{Video Taxonomy Prompt},
        colback=gray!5,
        colframe=gray!50,
        width=\textwidth, 
        boxrule=0.5pt,
        fonttitle=\bfseries,
        enhanced,
        left=5pt, right=5pt, top=5pt, bottom=5pt 
    ]
    \begin{Verbatim}[
    fontfamily=tt,
    fontsize=\tiny,
    breaklines=true,
    breakanywhere=false,
    breaksymbolleft={},  % 【关键】清除换行后下一行开头的符号
    breaksymbol={},      % 【关键】清除行尾的符号
    breaksymbolsep=0pt]
You are a video understanding expert. Analyze the following sequence of video frames in chronological order and classify the video into up to 3 most likely subcategories from the predefined taxonomy. For each predicted subcategory, provide a confidence probability (0.0–1.0). If the video cannot be confidently classified into any defined category, include "Others" with an appropriate probability. Output only a JSON dictionary in the format: {"subcategory1": probability1, "subcategory2": probability2, ...}. Probabilities do not need to sum to 1.0. Do not include any explanation or formatting.

Taxonomy Structure:
**Knowledge**
- **Nature**: Content about the natural world on Earth, including ecosystems, wildlife, environmental processes, and conservation efforts. Focuses on non-human-driven phenomena and the interdependence of living organisms and their habitats.
- **Science**: Systematic knowledge of the physical and technological world, including fundamental principles in physics, chemistry, biology, astronomy, and engineering. Covers how things work, from subatomic particles to space exploration, and the development of technologies such as AI and robotics.
- **Health**: Knowledge about the human body, medical science, disease prevention, mental well-being, and public health. Emphasizes evidence-based understanding of health conditions, treatments, and lifestyle impacts on physical and psychological wellness.
- **History**: Documented understanding of past human events, civilizations, conflicts, discoveries, and cultural developments. Based on historical records, archaeological findings, and scholarly analysis of how societies have evolved over time.
- **Society**: Insights into human social structures, behaviors, institutions, and collective thought. Includes economics, psychology, education, ethics, philosophy, and the study of how individuals and groups interact within cultural and organizational contexts.
**Everyday**
- **Human Bonds**: Authentic moments of connection and coexistence with family, friends, partners, or acquaintances, emphasizing emotional intimacy, shared experiences, and the warmth of human relationships.
- **Animal Companions**: Daily life and emotional bonding between humans and their animal companions, highlighting care, spontaneity, and the unique non-verbal intimacy shared across species.
- **Personal Life**: Recordings of an individual’s daily existence in solitude, encompassing routines, habits, reflections, emotions, domestic activities, and atmospheric moments. Focuses on how a person experiences, manages, and expresses their life without interaction with people or pets. This includes personal journeys, functional tasks, and contemplative states, all centered on the self as the sole subject.
**Creativity**
- **Movie & Show**: Fictional or dramatic videos that tell a story, including movies, TV series, web dramas, and short films. Typically feature actors, scripts, and narrative structure.
- **Dance & Performance**: Choreographed or expressive performances centered on movement, including original dance routines, dance covers, stage shows, spoken word poetry, and artistic recitations. Emphasizes physical expression, rhythm, and emotional delivery.
- **Music & Singing**: Original or performed musical works, including official music videos, song releases, vocal covers, instrumental performances, and creative audio-visual compositions. Focuses on auditory artistry and musical expression.
- **Animation**: Animated works created through 2D, 3D, stop-motion, or digital techniques, including short films, creative explainers, and experimental visual stories. Emphasizes visual imagination and motion design.
- **Comedy Sketch**: Short, scripted humorous videos designed to entertain, including parodies, satirical scenes, original comedy skits, and creative spoofs. Often feature exaggerated characters and comedic timing.
- **Game**: Creative content made within or about games, such as custom maps, mods, character designs, in-game art projects, or narrative-driven gameplay. Emphasizes originality, design, and virtual world-building.
- **Sports**: Content centered on athletic competitions and physical performance, including live events, highlights, athlete stories, news, and expert analysis. Emphasizes the drama, skill, and emotional intensity of sports as a form of visual and emotional entertainment.
**Civics**
- **News**: Reporting on recent, impactful public events such as natural disasters, accidents, conflicts, or major societal incidents. Focuses on what happened, where, and when, with emphasis on timeliness and factual accuracy.
- **Social Issues**: Coverage of ongoing societal challenges and public debates, such as education inequality, mental health awareness, housing affordability, gender rights, racial justice, and environmental policy. Focuses on current events, stakeholder perspectives, and civic discourse.
- **Civic Action**: Recordings of collective efforts to address social or environmental issues, such as protests, volunteer work, humanitarian aid, and community organizing. Highlights public participation and social change.
**Function**
- **How-To**: Step-by-step instructions for completing practical tasks in daily life, work, or learning—excluding cooking—such as repairing a device, using software, organizing space, crafting, or performing a physical skill. Covers both short-term actions and repeatable routines, with a focus on actionable guidance and immediate application.
- **Cooking**: Step-by-step instructions for preparing meals, dishes, or beverages, including recipe demonstrations, cooking techniques, meal prep, and kitchen tips. Focuses on food creation, flavor development, and practical kitchen skills.
- **Buyer's Guide**: Content that helps viewers decide what to buy, including product reviews, comparisons, recommendations, unboxing, and live commerce. Emphasizes real-world usage, value assessment, and decision support.
- **Travel Planning**: Guides for designing a trip, including itinerary creation, budgeting, transportation, accommodation, and visa planning. Helps viewers prepare for travel with practical, organized advice.
- **Life Guide**: Guides that help viewers design sustainable, personalized life systems, such as minimalism, daily routines, or personal workflows. Focuses on the philosophy, structure, and long-term optimization of everyday living—beyond step-by-step instructions.
- **Functional Recordings**: Videos recorded for practical purposes, such as screen recordings, surveillance, dashcams, meeting logs, or training replays. Not intended for entertainment or artistic expression.
    \end{Verbatim}
    \end{tcolorbox}
    \vspace{-4mm}
    \caption{Video Taxonomy Prompt}
    \label{fig:vid_tax_prompt}
\end{figure*}

\begin{figure*}[t] 
    \centering
    \begin{tcolorbox}[
        title=\textbf{QA Generation Prompt},
        colback=gray!5,
        colframe=gray!50,
        width=\textwidth, 
        boxrule=0.5pt,
        fonttitle=\bfseries,
        enhanced,
        left=5pt, right=5pt, top=5pt, bottom=5pt 
    ]
    \begin{Verbatim}[
    fontfamily=tt,
    fontsize=\tiny,
    breaklines=true,
    breakanywhere=false,
    breaksymbolleft={},  % 【关键】清除换行后下一行开头的符号
    breaksymbol={},      % 【关键】清除行尾的符号
    breaksymbolsep=0pt]
You are an expert AI assistant specializing in video analysis and question-answering (QA) generation. Your task is to analyze a sequence of video frames and generate a diverse set of high-quality, challenging, and **independently answerable** question-answering pairs based on the visual content alone.

### **Crucial Constraint: Independent Answerability**

This is the most important rule. A question is **independently answerable** if its question stem is a complete, self-contained instruction that can be answered by observing the video *without* seeing the options first. The options should only serve to test if the correct answer was found, not to help understand the question itself.
**Allowed Question Patterns (DOs):**
*   **Perceptual-Factual (Type I):** Questions with objective, verifiable answers found directly in the video.
    *   *Examples:* "What color is the car?", "How many people are in the room?", "What is the immediate result of the person flipping the switch?", "What is written on the sign?", "At 0:45, what is the woman doing?"
*   **Interpretive-Inferential (Type II):** Questions requiring reasoning, but where the answer is a logical conclusion strongly supported by visual evidence in the video.
    *   *Examples:* "What is the person's primary goal?", "How would you describe the person's emotional state?", "What event is most likely being prepared for?", "What most likely happened just before this scene began?"
**Forbidden Question Patterns (DON'Ts):**
*   **Option-Dependent Questions:** The question relies on the options to be understood.
    *   *Forbidden Examples:* "Which of the following best describes...", "Which of these events happened last?", "Which statement is false?"
*   **Counterfactual Questions:** The question asks about something that did *not* happen.
    *   *Forbidden Examples:* "What would have happened if the person had turned left?", "What if the phone had not rung?"
### **Input:**
The input will be a list of video frames, representing a silent video clip.
### **Output Requirements:**
You must generate a JSON formatted `list of dicts`. Each dictionary represents a single QA pair and must contain the following four keys:
-   `"category"`: The corresponding **category** from the following taxonomy (e.g., "Entity", "Relational Reasoning").
-   `"question"`: The question about the video content. The question **must strictly adhere to the 'Independent Answerability' constraint defined above**.
-   `"options"`: A list of 5 unique strings representing plausible answers. The options should be plausible and create meaningful difficulty (i.e., they should be good "distractors").
-   `"answer"`: The single correct answer, which **must** be one of the strings from the `"options"` list.
### **Coverage Mandate:**
The final output list **must contain at least TEN QA pairs for each** of the categories defined below.
---

### **Category Definitions and Examples**

Below are the definitions for each category and reference examples to guide your generation.
**1. Category: `Entity`**
   - **Definition:** Assesses the model's ability to identify and recognize the presence of people, animals, objects, text or symbol.
   - **Examples:** [... Demos omitted for brevity ...]
**2. Category: `Attribute`**
   - **Definition:** Assesses the ability to identify visual properties (color, shape), count entities, or determine an entity's state (open/closed, on/off).
   - **Examples:** [... Demos omitted for brevity ...]
**3. Category: `Action`**
   - **Definition:** Assesses the ability to identify a simple action performed by a single entity or a complex interaction involving multiple entities.
   - **Examples:** [... Demos omitted for brevity ...]
**4. Category: `Event`**
   - **Definition:** Assesses the ability to recognize the overall, composite activity or situation depicted in the video.
   - **Examples:** [... Demos omitted for brevity ...]
**5. Category: `Setting`**
   - **Definition:** Assesses the ability to identify the overall environment or infer the location and time based on visual cues.
   - **Examples:** [... Demos omitted for brevity ...]
**6. Category: `Relational Reasoning`**
   - **Definition:** Assesses the ability to infer spatial, temporal, or comparative relationships between entities or actions.
   - **Examples:** [... Demos omitted for brevity ...]
**7. Category: `Causal Reasoning`**
   - **Definition:** Assesses the ability to infer causes for events (why?), immediate effects (what results?), or alternative outcomes (what if?).
   - **Examples:** [... Demos omitted for brevity ...]
**8. Category: `Interpretive Reasoning`**
   - **Definition:** Assesses the ability to understand the plot, infer the underlying abstract theme, or find the symbolic meaning of the content.
   - **Examples:** [... Demos omitted for brevity ...]
**9. Category: `Intent & Emotion Reasoning`**
   - **Definition:** Assesses the ability to infer a person's goals, motivations, emotional state, or the specific purpose behind their actions.
   - **Examples:** [... Demos omitted for brevity ...]
**10. Category: `Grounding Reasoning`**
   - **Definition:** Assesses the ability to locate or verify content based on a text, temporal, or spatial reference.
   - **Examples:** [... Demos omitted for brevity ...]

Now, based on the following list of video frames, generate the QA pairs according to all the rules specified above.
    \end{Verbatim}
    \end{tcolorbox}
    \vspace{-4mm}
    \caption{QA Generation Prompt}
    \label{fig:qa_gen_prompt}
\end{figure*}

\begin{figure*}[t] 
    \centering
    \begin{tcolorbox}[
        title=\textbf{Blind Solvability Check Prompt},
        colback=gray!5,
        colframe=gray!50,
        width=\textwidth, 
        boxrule=0.5pt,
        fonttitle=\bfseries,
        enhanced,
        left=5pt, right=5pt, top=5pt, bottom=5pt 
    ]
    \begin{Verbatim}[
    fontfamily=tt,
    fontsize=\tiny,
    breaklines=true,
    breakanywhere=false,
    breaksymbolleft={},  % 【关键】清除换行后下一行开头的符号
    breaksymbol={},      % 【关键】清除行尾的符号
    breaksymbolsep=0pt]
You are a precise answering bot. Your only task is to solve the multiple-choice question provided below.
# Rules:
1.  Your response MUST be a single capital letter corresponding to the correct answer (e.g., A, B, C, D, or E).
2.  You are STRICTLY FORBIDDEN from providing any explanation, analysis, punctuation, or any text other than the single letter.
# Question & Options:
Question:
{question}
Options:
{options}
# Correct Option Letter:
    \end{Verbatim}
    \label{fig:filter_prompt}
    \end{tcolorbox}

    \begin{tcolorbox}[
        title=\textbf{Question Deduplication Prompt},
        colback=gray!5,
        colframe=gray!50,
        width=\textwidth, 
        boxrule=0.5pt,
        fonttitle=\bfseries,
        enhanced,
        left=5pt, right=5pt, top=5pt, bottom=5pt 
    ]
    \begin{Verbatim}[
    fontfamily=tt,
    fontsize=\tiny,
    breaklines=true,
    breakanywhere=false,
    breaksymbolleft={},  % 【关键】清除换行后下一行开头的符号
    breaksymbol={},      % 【关键】清除行尾的符号
    breaksymbolsep=0pt]
# Role
You are an expert in Natural Language Understanding (NLU) and data clustering.
# Task
Your task is to perform a two-level semantic clustering on a list of QA pairs. The process is as follows:
1.  **First Level Grouping**: Group all QA pairs that are semantically consistent at the **Question** level.
2.  **Second Level Grouping**: Within each question group, further group the QA pairs that are semantically consistent at the **Answer** level.
Finally, you must output a nested list of indices that reflects this hierarchical clustering structure.
# Input Format
The input is a JSON list of objects. Each object contains three fields: `index` (an integer starting from 0), `question` (a string), and `answer` (a string).
Example:
```json
[
  {"index": 0, "question": "...", "answer": "..."},
  {"index": 1, "question": "...", "answer": "..."},
  ...
]
```
# Output Format
The output must be a JSON-formatted nested list of integers (`List[List[List[int]]]`).
- The **innermost list** (`List[int]`): Contains the `index` of QA pairs where both the question and answer are semantically consistent.
- The **middle list** (`List[List[int]]`): Represents a group of QA pairs with semantically consistent questions.
- The **outermost list** (`List[List[List[int]]]`): The final result containing all groups.
# Core Rules and Constraints
1.  **Completeness**: Every `index` from the input must appear in the output exactly once.
2.  **Atomicity**: If a QA pair is semantically unique (its question and answer do not match any others), it will form its own singleton group, e.g., `[[[index]]]`.
3.  **Strict Format**: Your final response must only be the required JSON nested list. Do not include any explanations or surrounding text.
# Semantic Consistency Guidelines (Relaxed Standard)
**1. Question Consistency (Relaxed):**
Questions are considered consistent if they ask about the **same core topic or intent**, even if the phrasing, scope, or specific focus varies.
- **CONSIDER CONSISTENT**:
    - "What is the man in the video wearing?" vs. "Can you describe the person's attire?"
    - "What is the main topic of the video?" vs. "Give a summary of this video."
- **DO NOT CONSIDER CONSISTENT**:
    - "What is the character doing?" vs. "What is the character wearing?" (Action vs. Appearance)
    - "What's the music in the background?" vs. "Where was the video filmed?" (Audio vs. Location)
**2. Answer Consistency (Relaxed):**
Answers are considered consistent if they convey the **same core information or conclusion**. Minor differences in wording, detail level, or filler words should be ignored.
- **CONSIDER CONSISTENT**:
    - "He is wearing a blue jacket." vs. "A blue jacket."
    - "The video teaches how to bake a cake." vs. "It's a tutorial about cake baking."
- **DO NOT CONSIDER CONSISTENT**:
    - "He is wearing a blue jacket." vs. "He is wearing a red shirt." (Different core information)
    - "Yes, the car is speeding." vs. "It is uncertain if the car is speeding." (Contradictory conclusions)
Now, based on all the rules and guidelines above, process the following list of QA pairs.
    \end{Verbatim}
    \label{fig:dedup_prompt}
    \end{tcolorbox}
    \vspace{-4mm}
    \caption{Blind Solvability Check and Question Deduplication Prompt}
    \label{fig:to_dedep}
\end{figure*}

\begin{figure*}[t] 
    \centering
    \begin{tcolorbox}[
        title=\textbf{Multiple-Choice Question Answering Prompt},
        colback=gray!5,
        colframe=gray!50,
        width=\textwidth, 
        boxrule=0.5pt,
        fonttitle=\bfseries,
        enhanced,
        left=5pt, right=5pt, top=5pt, bottom=5pt 
    ]
    \begin{Verbatim}[
    fontfamily=tt,
    fontsize=\tiny,
    breaklines=true,
    breakanywhere=false,
    breaksymbolleft={},  % 【关键】清除换行后下一行开头的符号
    breaksymbol={},      % 【关键】清除行尾的符号
    breaksymbolsep=0pt]
You are an intelligent video understanding assistant.
Select the best answer to the following multiple-choice question based on the video caption. Respond with only the letter ({option_id_str}) of the correct option.
**Video Caption**
{caption}
**Question**
{question}
**Options**
{options}
Answer the question with {option_id_str}.
    \end{Verbatim}
    \label{fig:choice_prompt}
    \end{tcolorbox}

    \begin{tcolorbox}[
        title=\textbf{VLLM-as-a-Judge Prompt},
        colback=gray!5,
        colframe=gray!50,
        width=\textwidth, 
        boxrule=0.5pt,
        fonttitle=\bfseries,
        enhanced,
        left=5pt, right=5pt, top=5pt, bottom=5pt 
    ]
    \begin{Verbatim}[
    fontfamily=tt,
    fontsize=\tiny,
    breaklines=true,
    breakanywhere=false,
    breaksymbolleft={},  % 【关键】清除换行后下一行开头的符号
    breaksymbol={},      % 【关键】清除行尾的符号
    breaksymbolsep=0pt]
You are an expert Video Understanding Evaluator. Your task is to assess the quality of a generated video caption by strictly comparing it against the provided video file.
You must evaluate the caption from three distinct perspectives using a strict 1-to-5 scale. You will first analyze the video content to establish the ground truth, and then assign scores based on the detailed rubrics provided below.
**Perspectives:**
1. Trustworthiness (Precision): Focuses purely on factuality. Penalizes hallucinations and incorrect information.
2. Coverage (Recall): Focuses on completeness. Penalizes missing salient events, objects, or context.
3. Overall Quality: A holistic assessment of the caption's descriptive value.
### Task
Watch the video carefully and evaluate the Candidate Caption.
**Candidate Caption:**
"{candidate_caption}"
### Evaluation Steps
Please think step-by-step:
**Step 1: Video Analysis (Internal Ground Truth)**
- Identify the main subject, the primary action/event, and the setting.
- Note the chronological sequence of events.
**Step 2: Scoring via Rubrics**
Assign an integer score (0-5) for each dimension based strictly on the following definitions:
#### 1. Trustworthiness (Precision) Rubric
*Does the caption contain false information?*
- **5 (Perfect):** No factual errors. Every detail (action, object, color, count) is visually confirmed by the video.
- **4 (High):** Factually correct, but may contain slight ambiguities or imprecisions that are not explicitly wrong (e.g., calling a "spaniel" just a "dog").
- **3 (Moderate):** Mostly correct, but contains **one minor hallucination** (e.g., wrong color of a shirt, or a minor background detail is wrong).
- **2 (Low):** Contains **multiple minor errors** or **one significant error** (e.g., misinterpreting a secondary action).
- **1 (Very Low):** Contains **severe hallucinations**. Describes a main action or object that is not present in the video at all.
#### 2. Coverage (Recall) Rubric
*Does the caption miss important information?*
- **5 (Perfect):** Covers the main event, key details, temporal progression, and setting. Nothing important is left out.
- **4 (High):** Covers the main event and setting, but misses **trivial/minor visual details** (e.g., background objects).
- **3 (Moderate):** Covers the gist of the video, but misses **one salient detail** or part of the context (e.g., mentions the action but misses the object being acted upon).
- **2 (Low):** Misses the **primary action** or the **climax** of the video. Focuses only on the setup or the aftermath.
- **1 (Very Low):** Captures only tangential or irrelevant background details. Misses the main story entirely.
#### 3. Overall Quality Rubric
*How useful is this caption to a blind user?*
- **5 (Excellent):** Accurate, complete, and fluent. Perfectly describes the video.
- **4 (Good):** Very useful, with only negligible flaws in detail or fluency.
- **3 (Fair):** Acceptable. Convey the main idea but has noticeable flaws in accuracy or completeness.
- **2 (Poor):** Misleading or confusing. Requires the user to guess what happened.
- **1 (Bad):** Not useful. Contains high levels of noise or error.
### Output Format
Provide the output in strict JSON format only:
{
  "analysis": "Brief analysis of video vs. caption...",
  "scores": {
    "trustworthiness": <int 1-5>,
    "coverage": <int 1-5>,
    "overall": <int 1-5>
  }
}
    \end{Verbatim}
    \label{fig:eval_prompt}
    \end{tcolorbox}
    \vspace{-4mm}
    \caption{Multiple-Choice Question Answering and LLM Grader Prompt}
    \label{fig:grade_prompt}
\end{figure*}

\begin{figure*}[t] 
    \centering
    \begin{tcolorbox}[
        title=\textbf{Detailed Caption Prompt},
        colback=gray!5,
        colframe=gray!50,
        width=\textwidth, 
        boxrule=0.5pt,
        fonttitle=\bfseries,
        enhanced,
        left=5pt, right=5pt, top=5pt, bottom=5pt 
    ]
    \begin{Verbatim}[
    fontfamily=tt,
    fontsize=\tiny,
    breaklines=true,
    breakanywhere=false,
    breaksymbolleft={},  % 【关键】清除换行后下一行开头的符号
    breaksymbol={},      % 【关键】清除行尾的符号
    breaksymbolsep=0pt]
**You are a world-class Visual Intelligence Analyst.** Your mission is to deconstruct a sequence of chronologically ordered images (video clips) into a definitive, structured report. You have **NO audio information**. Your analysis must be purely visual, objective, and meticulously detailed, serving as a complete knowledge base to answer any possible subsequent question.

You **must** generate a report that strictly adheres to the following hierarchical structure and fills every field with as much detail as possible.

---

### **1. Executive Summary & Scene Classification**
*   **Core Narrative Synopsis**: (A concise, one-to-two-sentence summary encapsulating the primary action, subjects, and outcome.)
*   **Inferred Genre/Context**: (e.g., Cooking tutorial, product review, documentary segment, home video, animated short, security footage.)

### **2. Perception Analysis: The Observable Reality**
*   **2.1. Entity & Attribute Identification**:
    *   **Characters/People**:
        *   `[Person A]`:
            *   **Visual Properties**: Describe appearance (gender, age est., hair, ethnicity), clothing (type, color, style), and accessories.
            *   **Quantity**: (e.g., "One person initially, a second person enters at time 00:00:45.")
    *   **Key Objects**:
        *   `[Object A]`:
            *   **Visual Properties**: Describe its type, color, material, shape, and any distinct features.
            *   **Quantity**: Note the count of similar objects (e.g., "Three blue cups on the table.").
            *   **State & State Changes**: Describe its condition or status and note any changes (e.g., "Initially closed, opened at time 00:00:30," "Power light is on," "Appears damaged.").
    *   **Animals**:
        *   `[Animal A]`: Describe species, breed, color, size.
*   **2.2. Environment & Setting Analysis**:
    *   **Location Type**: (e.g., Indoor kitchen, outdoor public park, office cubicle, car interior.)
    *   **Ambient Details**: Note key background elements, furniture, weather conditions, and general state (e.g., "tidy," "cluttered").
    *   **Inferred Time of Day**: (e.g., "Bright daylight due to harsh shadows," "Dusk inferred from warm, low light," "Night, lit by artificial sources.")
*   **2.3. On-Screen Text & Graphics**:
    *   `[Text/Graphic 1]`: Transcribe the text/logo and note its location and the time ranges in which it is visible, e.g. 00:00:15-00:00:30.

### **3. Reasoning Analysis: Connecting the Dots**
*   **3.1. Chronological & Causal Reconstruction**:
    *   `Time Segment [e.g., 00:00:00-00:00:10]`:
        *   **Atomic Actions**: Describe actions by single entities (e.g., "Person A picks up the red ball.").
        *   **Interactions**: Describe actions between entities (e.g., "Person A hands the ball to Person B.").
        *   **Causal Links**: If an action directly causes a result, state it (e.g., "**Because** the ball was thrown, the window broke.").
    *   `Time Segment [e.g., 00:00:10-00:00:20]`: (Repeat the structure above.)
*   **3.2. Relational Analysis**:
    *   **Spatial Relations**: Throughout the sequence, describe the key relative positions (e.g., "The cat is sleeping **under** the table," "At 00:00:50, Person A moves to stand **behind** Person B.").
    *   **Temporal Relations**: Use clear sequential language (e.g., "The phone rings **before** she opens the book," "**While** he was cooking, the dog entered the room.").
    *   **Comparative Observations**: Note any explicit or implicit comparisons (e.g., "The second car is moving **faster than** the first," "Box A is visibly **larger than** Box B.").
*   **3.3. Interpretive & Social Analysis**:
    *   **Inferred Emotions & Intent**:
        *   `[Person A]`: (e.g., "Appears focused and determined, likely intending to complete the puzzle," "Facial expression shifts from neutral to surprised at 00:01:00.").
    *   **Plot & Thematic Reasoning**: Describe the overall story, moral, or abstract message being conveyed.
    *   **Social & Normative Context**: Describe the social dynamics (e.g., "formal interview," "casual conversation," "teacher-student interaction"). Note any actions that align with or deviate from common social norms.

### **4. Prediction & Extrapolation Analysis**
*   **4.1. Immediate Next Action (Short-term Prediction)**: Based on the final timestamp, what is the single most likely action to occur in the next 1-3 seconds?
*   **4.2. Plausible Outcome (Long-term Prediction)**: What is the probable final outcome or resolution of the entire event shown? (e.g., "The meal will be successfully prepared," "The argument will be resolved.").
*   **4.3. Pre-Condition Inference (Past Inference)**: What events or conditions likely occurred *before* the video started to create the initial scene?
*   **4.4. Counterfactual Hypothesis (What-If Scenario)**: Propose one key "what-if" scenario. (e.g., "If Person A had not dropped the keys, they would have likely caught the bus on time.").

### **5. Meta-Analysis**
*   **Cinematography & Style**: (e.g., "Handheld camera with shaky movement," "Static tripod shot," "High-contrast, cinematic lighting," "Minimalist aesthetic.").
    \end{Verbatim}
    \end{tcolorbox}
    \vspace{-4mm}
    \caption{Detailed Caption Prompt}
    \label{fig:detail_prompt}
\end{figure*}

\section{Annotation Details}
\label{app:annotation_details}

\begin{figure*}[t]
    \centering
    \includegraphics[width=\textwidth]{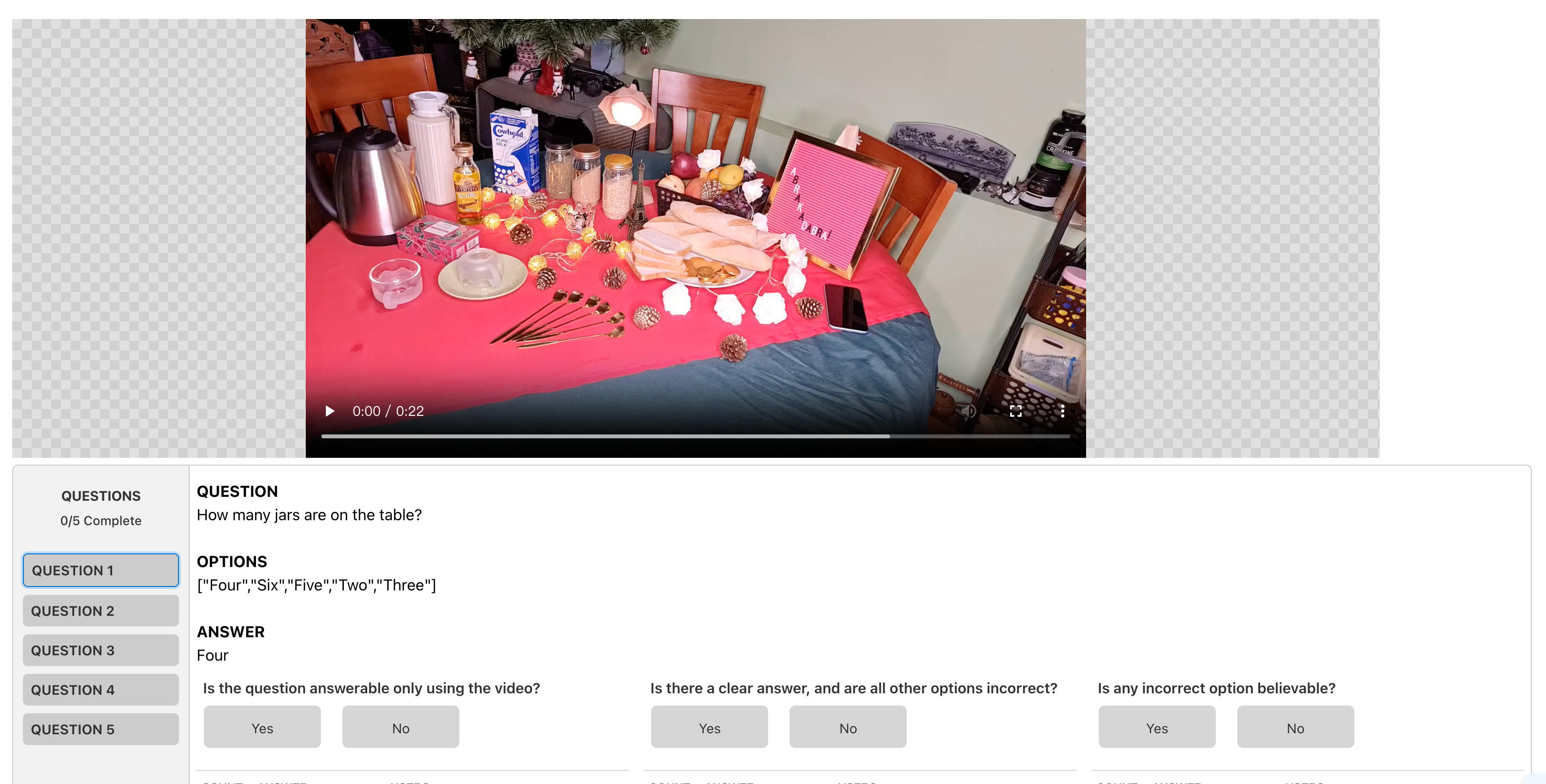}
    \vspace{-2mm} 
    \caption{The screenshot of Human Verification system.}
    \label{fig:screenshot}
    \vspace{-4mm} 
\end{figure*}

We recruited annotators through our internal data annotation platform.
Our annotation workforce exhibits a diverse international background, roughly evenly split into four groups. 
India, Singapore, and China each contribute approximately 25\% of the personnel, while the final cohort represents a mix of European and North American countries, including the United States, Spain, and Ireland.
To ensure high-quality text generation and comprehension, we enforced a strict prerequisite of proficiency in written English.
The annotation interface used by the workers is illustrated in Figure \ref{fig:screenshot}.



\label{sec:appendix}

\end{document}